\documentclass[10pt,twocolumn,letterpaper]{article}

\usepackage{cvpr}              % To produce the CAMERA-READY version
\usepackage{color, colortbl}

\newcommand{\worsetext}[1]{\textbf{\;({\color[rgb]{0.8, 0.2, 0.2}{#1}})}}
\newcommand{\ignoretext}[1]{\color[rgb]{0, 0, 0}{#1}}

\newcommand{\bslcell}[1]{\cellcolor[rgb]{1, 1, 1}{#1}}
\definecolor{highlight}{RGB}{236, 236, 236}

\newcommand{\ourname}[1]{UniHOPE}

\usepackage{dsfont}

\definecolor{cvprblue}{rgb}{0.21,0.49,0.74}
\usepackage[pagebackref,breaklinks,colorlinks,allcolors=cvprblue]{hyperref}
\usepackage[utf8]{inputenc} % allow utf-8 input
\usepackage[T1]{fontenc}    % use 8-bit T1 fonts
\usepackage{url}            % simple URL typesetting
\usepackage{booktabs}       % professional-quality tables
\usepackage{amsfonts}       % blackboard math symbols
\usepackage{nicefrac}       % compact symbols for 1/2, etc.
\usepackage{microtype}      % microtypography
\usepackage{xcolor}         % colors
\usepackage{graphicx}

\usepackage{color}
\usepackage{epsfig}
\usepackage{leftidx}
\usepackage{amsmath}
\usepackage{amssymb}
\usepackage{mathtools}
\usepackage{bbding}
\usepackage{booktabs}
\usepackage{multirow}
\usepackage{makecell}
\usepackage{enumitem}
\usepackage{dsfont}
\usepackage{array}
\usepackage{xcolor}

\usepackage{tikz}
\usepackage[accsupp]{axessibility}  % Improves PDF readability for those with disabilities.
\usetikzlibrary{tikzmark, arrows, arrows.meta}

\newcommand\blfootnote[1]{%
  \begingroup
  \renewcommand\thefootnote{}\footnote{#1}%
  \addtocounter{footnote}{-1}%
  \endgroup
}

\def\paperID{1934} % *** Enter the Paper ID here
\def\confName{CVPR}
\def\confYear{2025}

\title{\ourname{}: A Unified Approach for Hand-Only and Hand-Object Pose Estimation}

\author{
Yinqiao Wang$^{*}$ \hspace{0.5cm} 
Hao Xu$^{*}$ \hspace{0.5cm}
Pheng-Ann Heng \hspace{0.5cm}
Chi-Wing Fu \\
Department of Computer Science and Engineering\\
Institute of Medical Intelligence and XR\\
The Chinese University of Hong Kong\\
{\tt\small \{yqwang,xuhao,pheng,cwfu\}@cse.cuhk.edu.hk}
}

\begin{document}
\maketitle
\blfootnote{$^*$ Equal contribution.}

\vspace{-10mm}
\begin{abstract}
Estimating the 3D pose of hand and potential hand-held object from monocular images is a longstanding challenge. 
Yet, existing methods are specialized, focusing on either bare-hand or hand interacting with object.
No method can flexibly handle both scenarios and their performance degrades when applied to the other scenario. 
In this paper, we propose \ourname{}, a unified approach for general 3D hand-object pose estimation, flexibly adapting both scenarios. 
Technically, we design a grasp-aware feature fusion module to integrate hand-object features with an object switcher to dynamically control the hand-object pose estimation according to grasping status.
Further, to uplift the robustness of hand pose estimation regardless of object presence, we generate realistic de-occluded image pairs to train the model to learn object-induced hand occlusions, and formulate multi-level feature enhancement techniques for learning occlusion-invariant features. 
Extensive experiments on three commonly-used benchmarks demonstrate \ourname{}'s SOTA performance in addressing hand-only and hand-object scenarios.
Code will be released on \href{https://github.com/JoyboyWang/UniHOPE_Pytorch}{https://github.com/JoyboyWang/UniHOPE\_Pytorch}.
\end{abstract}

\vspace{-6mm}
\section{Introduction}
\label{sec1:intro}

\begin{figure}[t]
    \centering
    \includegraphics[width=0.99\linewidth]{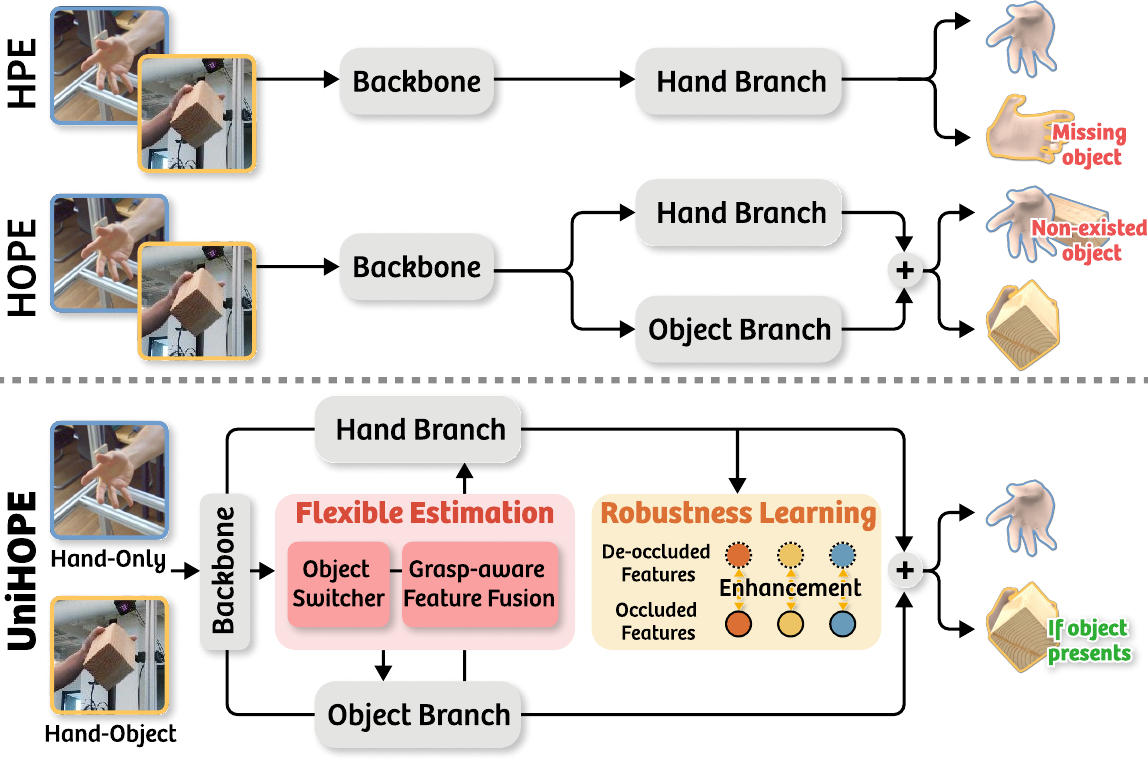}
    \vspace*{-2.5mm}
    \caption{Existing approaches (top) for 3D hand pose estimation are either Hand Pose Estimation (HPE), which predicts hand pose only, or Hand-Object Pose Estimation (HOPE), which assumes hand-held object. 
    Our novel \ourname{} approach (bottom) offers flexibility and robustness to handle both scenes in a unified manner.
    }
    \label{fig:teaser}
    \vspace*{-5mm}
\end{figure}
\begin{table*}[t]
    \centering
    \resizebox{\linewidth}{!}{
        \setlength\tabcolsep{12pt}
        \begin{tabular}{l|cc|cc|cc|cc}
        \toprule
        \multicolumn{1}{c|}{\multirow{2}{*}{\textbf{HPE}}} & \multicolumn{2}{c|}{\textit{Hand-Only Scene}} & \multicolumn{2}{c|}{\textit{Hand-Only $\rightarrow$ Hand-Object Scene}} & \multicolumn{2}{c|}{\textit{All $\rightarrow$ Hand-Only Scene}} & \multicolumn{2}{c}{\textit{All $\rightarrow$ Hand-Object Scene}} \\
        \cmidrule{2-9}
        & J-PE $\downarrow$ & V-PE $\downarrow$ & J-PE $\downarrow$ & V-PE $\downarrow$ & J-PE $\downarrow$ & V-PE $\downarrow$ & J-PE $\downarrow$ & V-PE $\downarrow$ \\
        \cmidrule{1-9}
        HandOccNet~\cite{park2022handoccnet} & \bslcell{12.98} & \bslcell{12.52} & 19.60 \worsetext{-6.62} & 18.95 \worsetext{-6.43} & 13.16 \worsetext{-0.18} & 12.70 \worsetext{-0.18} & \ignoretext{14.58} & \ignoretext{14.10} \\
        H2ONet~\cite{xu2023h2onet} & \bslcell{13.34} & \bslcell{13.13} & 21.98 \worsetext{-8.64} & 21.42 \worsetext{-8.30} & 14.14 \worsetext{-0.80} & 14.00 \worsetext{-0.87} & \ignoretext{15.20} & \ignoretext{15.03} \\
        SimpleHand~\cite{zhou2024simple} & \bslcell{14.05} & \bslcell{13.51} & 18.37 \worsetext{-4.32} & 17.54 \worsetext{-4.03} & 14.63 \worsetext{-0.58} & 13.96 \worsetext{-0.45} & \ignoretext{14.88} & \ignoretext{14.21} \\
        \midrule 
        \addlinespace[1mm]
        \midrule
        \multicolumn{1}{c|}{\multirow{2}{*}{\textbf{HOPE}}} & \multicolumn{2}{c|}{\textit{Hand-Object Scene}} & \multicolumn{2}{c|}{\textit{Hand-Object $\rightarrow$ Hand-Only Scene}} & \multicolumn{2}{c|}{\textit{All $\rightarrow$ Hand-Object Scene}} & \multicolumn{2}{c}{\textit{All $\rightarrow$ Hand-Only Scene}} \\
        \cmidrule{2-9}
        & J-PE $\downarrow$ & V-PE $\downarrow$ & J-PE $\downarrow$ & V-PE $\downarrow$ & J-PE $\downarrow$ & V-PE $\downarrow$ & J-PE $\downarrow$ & V-PE $\downarrow$ \\
        \cmidrule{1-9}
        Keypoint Trans.~\cite{hampali2022keypoint} & \bslcell{17.99} & \bslcell{17.57} & 25.10 \worsetext{-7.11} & 24.40 \worsetext{-6.83} & 18.79 \worsetext{-1.00} & 18.35 \worsetext{-0.78} & \ignoretext{19.75} & \ignoretext{19.26} \\
        HFL-Net~\cite{lin2023harmonious} & \bslcell{14.61} & \bslcell{14.13} & 19.39 \worsetext{-4.78} & 18.61 \worsetext{-4.48} & 14.77 \worsetext{-0.16} & 14.29 \worsetext{-0.16} & \ignoretext{13.61} & \ignoretext{13.10} \\
        \bottomrule
        \end{tabular}
        }
    \vspace{-3mm}
    \caption{
    Existing HPE methods trained on hand-only scene exhibit obvious performance degradation when testing on hand-object scene (1st \vs 2nd columns).
    Though training on all scenes (3rd \& 4th columns) helps to improve metrics on the hand-object scene (2nd \vs 4th columns), their original performance is adversely affected (1st \vs 3rd columns).
    The HOPE methods also exhibit a similar pattern (see bottom part).
    These results demonstrate the inabilities of the existing methods to flexibly handle hand-only and hand-object scenes altogether.
    }
    \vspace{-4mm}
    \label{tab:validation}
\end{table*}
Estimating the 3D pose of hand and potential hand-held objects from monocular images is a long-standing task with applications in VR/AR, human-computer interactions,~\etc. 

However, existing methods are divided.
As \cref{fig:teaser} illustrates, hand pose estimation (HPE) methods~\cite{moon2020i2l, cao2021reconstructing, park2022handoccnet, chen2022mobrecon, huang2023neural, xu2023h2onet, zhou2024simple} predict the 3D hand pose without considering the hand-held object.
Conversely, hand-object pose estimation (HOPE) methods~\cite{hasson2019learning, liu2021semi, hampali2022keypoint, wang2023interacting, lin2023harmonious} assume the presence of a hand-held object and perform object pose estimation with an extra object branch. Yet, they always make predictions even there is no object.
Neither approach offers the flexibility to consider both hand-only and hand-object scenarios.

\cref{tab:validation} provides a detailed analysis of the performance of state-of-the-art (SOTA) HPE methods~\cite{park2022handoccnet, xu2023h2onet, zhou2024simple} and HOPE methods~\cite{hampali2022keypoint, lin2023harmonious}. 
We observe an obvious performance degradation when these methods are applied across different scenes (see ``\emph{Hand-Only$\leftrightarrow$Hand-Object Scene}'') due to their task-specific designs.
Though training on all scenes helps, it negatively impacts their original task performance (see ``\emph{All$\rightarrow$Hand-Only/Hand-Object Scene}''), revealing their limited generalization capabilities. 
This observation motivates the need for a unified approach that can adapt effectively to both hand-only and hand-object scenes.

In this work, we present \ourname{}, the first method to unify HPE and HOPE by addressing (i) basic criteria: {\em adaptively switch between two scenes}; 
and (ii) advanced criteria: {\em robustly estimate hand pose regardless of object presence}. 

First, to meet the basic criteria, we propose that the hand pose must always be predicted regardless of whether the hand is grasping an object or not, while the object pose should be estimated only if the object is present.
Though a straightforward solution is to manually select existing SOTA HPE and HOPE methods according to the input scene, this approach is suboptimal, as switching between models leads to incoherent results and prevents joint optimization for one model to work on both scenes.
Importantly, we design an end-to-end method that dynamically controls object pose estimation through an internal object switcher by estimating the confidence of the grasping status, thereby 
promoting compatibility with various scenarios.
However, solely adopting existing network architectures along with our object switcher encounters another issue caused by the commonly-used hand-object information interaction structure~\cite{liu2021semi, tse2022collaborative, lin2023harmonious}.
When no object is present, extracting object features is unnecessary, so irrelevant object-to-hand feature transitions compromise hand pose estimation accuracy. 
To overcome this issue, we formulate a grasp-aware feature fusion module to utilize grasping confidence to select effective object features in hand-object feature fusion.

Second, the advanced criteria emphasize coherent and robust hand pose estimation, regardless of whether the hand is grasping an object or not. 
As hand-held objects frequently cause severe occlusions, it is essential to learn occlusion-invariant features to accurately recover hand poses. 
Ideally, the feature of a non-occluded hand serves as the optimal representation for an occluded hand with the same pose, as it simplifies the prediction difficulty. 
To facilitate this, we propose training the model by transferring knowledge from the corresponding non-occluded hands to the occluded ones. 
Due to the scarcity of such paired data, we innovatively leverage diffusion-based generative models to create realistic de-occluded hand images from the originally-occluded ones. 
Subsequently, we adopt multi-level feature enhancement techniques to help the network simulate occlusion-invariant features by utilizing information from the de-occluded hand images in a self-distillation framework. 

Our main contributions are summarized as follows:
\begin{itemize}
\item 
We demonstrate the necessity for a unified solution to hand-object pose estimation and propose \ourname{}, a novel approach to handle general hand-object scenarios.
\item 
We design an internal object switcher that provides flexibility across different scenes and formulate a grasp-aware feature fusion module to adaptively utilize the effective object information based on grasping status.
\item 
We propose an occlusion-invariant feature learning strategy for robustness, first using a generative de-occluder to prepare paired de-occluded hand images and then applying feature enhancement at multiple levels. 
\item 
Extensive experiments on three widely-used datasets in our unified setting show the SOTA performance of \ourname{}.
\end{itemize}

\begin{figure*}[t]
    \centering
    \includegraphics[width=\linewidth]{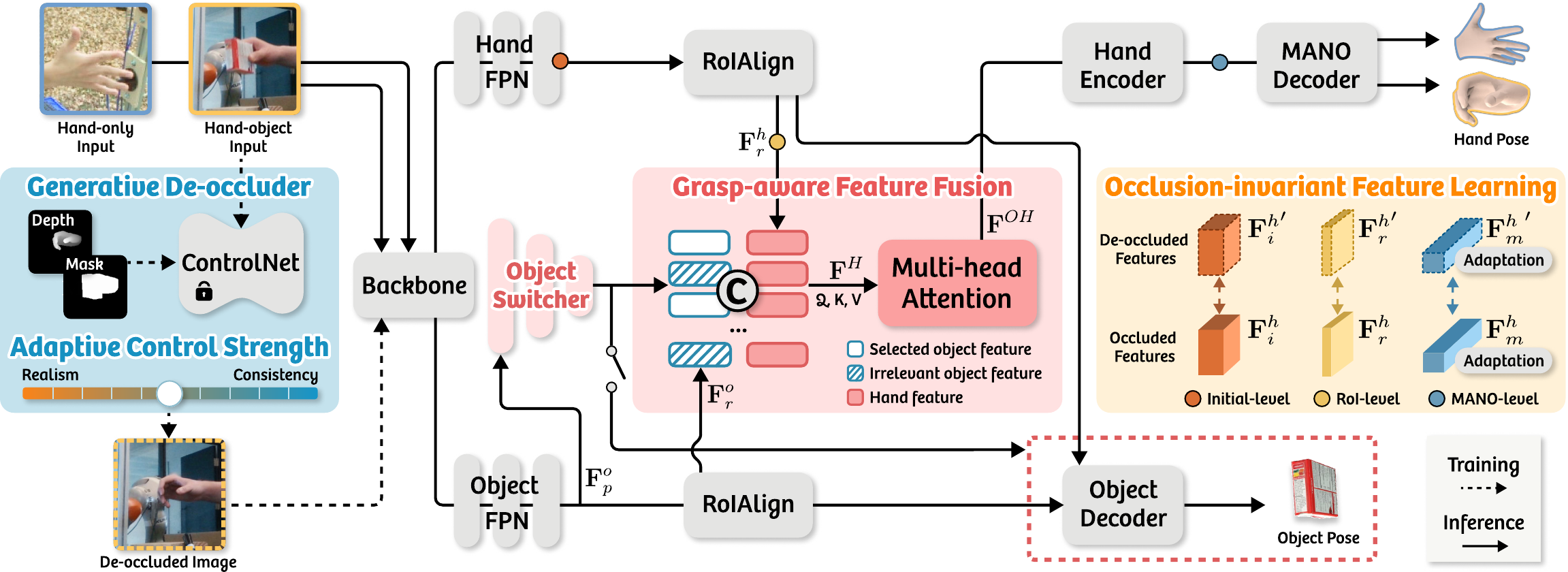}
    \vspace{-7mm}
     \caption{Our \ourname{} framework.
    (i) We first de-occlude hand images occluded by objects to form pairs, conditioned on the depth map and hand-object mask, with adaptive selection of control strength to produce high-quality samples;
    (ii) to accommodate both hand-only and hand-object scenes, our object switcher dynamically controls the object output by predicting grasping status, which guides the feature fusion module to eliminate irrelevant object features; and
    (iii) to robustly estimate hand pose, our multi-level feature enhancement techniques utilize paired data to learn occlusion-invariant hand features.
    }
    \label{fig:pipeline}
    \vspace{-3mm}
\end{figure*}

\section{Related Work}
\label{sec2:rw}
\paragraph{Monocular 3D Hand Pose Estimation (HPE).}
Most methods formulate HPE by regressing MANO coefficients~\cite{zhang2019end, zhou2020monocular, yang2020bihand, zhang2021hand, boukhayma20193d, baek2019pushing, zimmermann2019freihand, chen2021model, zhao2021travelnet, baek2020weakly, zhang2021interacting}. 
Other common representations include voxels~\cite{iqbal2018hand, moon2020i2l, moon2020interhand2, yang2021semihand}, implicit functions~\cite{huang2023neural}, and meshes~\cite{kulon2020weakly, chen2021camera, chen2022mobrecon, xu2023h2onet, zhou2024simple}.
While they achieve superior performance in predicting hand poses, they do not account for object pose estimation during hand-object interactions, which is critical for practical applications.

\vspace{-5mm}
\paragraph{Monocular 3D Hand-Object Pose Estimation (HOPE).}
To jointly estimate the hand and object poses simultaneously, recent works can be categorized into two main streams:
(i) Template-free methods reconstruct objects without knowing their 3D models.
Hasson~\etal~\cite{hasson2019learning} recover the object mesh from a deformed icosphere. 
Tse~\etal~\cite{tse2022collaborative} propose to optimize hand and object meshes iteratively.
More recent works leverage implicit~\cite{chen2022alignsdf, ye2022s, huang2022reconstructing, chen2023gsdf, ye2024g} or neural fields~\cite{choi2024handnerf} to represent the hand and object; yet, these methods often struggle to accurately model unseen objects due to limited data prior. 
(ii) Template-based methods assume the 3D object model is known, focusing on regressing its pose.
Liu~\etal~\cite{liu2021semi} proposes a semi-supervised learning approach.
Lin~\etal~\cite{lin2023harmonious} design a dual-branch backbone to leverage mutual hand-object information. 
Though these methods produce promising results, they focus on the hand-object scene, lacking the flexibility to handle the hand-only scenario.

\vspace{-5mm}
\paragraph{Hand-Object Image Synthesis.}
Since we generate paired de-occluded hand images for unified scenarios, we also review methods for hand-object image synthesis.
Rendering-based methods~\cite{hasson2019learning, corona2020ganhand, yang2022artiboost} use common tools~\cite{pyrender, maya, blender} to render images as augmented data for HPE and HOPE tasks.
With the advance of generative models~\cite{goodfellow2020generative, sohl2015deep, ho2020denoising, song2020denoising, wang2022zero, rombach2022high},
recent methods generate more realistic data for various purposes, including image generation~\cite{hu2022hand, narasimhaswamy2024handiffuser, zhang2024hoidiffusion}, object-prior guidance~\cite{ye2023affordance}, hand re-malformation~\cite{lu2024handrefiner}, and training data augmentation~\cite{xu2024handbooster, park2024attentionhand}.
In this work, we propose to generate paired de-occluded hand images to facilitate learning occlusion-invariant hand features, thereby improving robustness to handle both hand-only and hand-object scenes. 

\section{Method}
\label{sec3:method}
\subsection{Overview}
Distinct from previous studies, we address a more general scenario,~\ie, the model always predicts hand pose, regardless of whether the hand is grabbing an object or not.
If an object is present, our model also estimates its pose.

\cref{fig:pipeline} shows our \ourname{} pipeline.
First, we propose to dynamically estimate hand-object pose in an end-to-end manner (see the red section in \cref{fig:pipeline}), where our object switcher flexibly controls the output and our grasp-aware feature fusion module integrates grasp-relevant object information (\cref{sec3.2:feature_fusion}).
Next, to improve robustness against occlusions, we first design our generative de-occluder to prepare high-quality paired data by adaptively adjusting control strengths (see blue section), which is then used to learn occlusion-invariant features through our multi-level feature enhancement techniques (see yellow section, \cref{sec3.3:feature_enhancement}). 
Finally, we detail the loss functions used in our approach in~\cref{sec3.4:loss}.
% 

%%%%%%%%%%%%%%%%%%%%%%%%%%%%%%%%%%%%%%%%%%
\vspace{-2mm}
\subsection{Dynamic Hand-Object Pose Estimation}
\label{sec3.2:feature_fusion}
To accommodate both HPE and HOPE, it is essential to consistently predict hand poses while estimating object poses only when an object is present.
Directly combining existing HPE and HOPE methods is insufficient due to the incoherence introduced by model switching and the lack of joint optimization.
In this section, we present our end-to-end dynamic hand-object pose estimation approach. 
To flexibly control the object output, we introduce an object switcher that predicts grasping status, trained with automatically-generated labels.
Further, we propose a grasp-aware feature fusion module guided by the grasping status to prevent object-to-hand irrelevant feature transitions when no object is present. 

\vspace{-4mm}
\paragraph{Grasping Label Preparation.}
To support model training, we automatically prepare grasping status labels, as existing datasets~\cite{chao2021dexycb, hampali2020honnotate} do not provide this information. 
Following~\cite{xu2024handbooster}, we compute the isotropic Relative Rotation Error (RRE) and Relative Translation Error (RTE) between the object poses in the initial and current frames:
\begin{equation}
\small
    \mathrm{RRE}\!=\!\arccos(\frac{\mathrm{trace}({\mathbf{\xi}_{R}^t}^{\top} \mathbf{\xi}_{R}^{0} \!-\! 1)}{2}),\; \mathrm{RTE}\!=\!||{\mathbf{\xi}_{T}^{t}-\mathbf{\xi}_{T}^{0}}||_2,
\end{equation}
where $\mathbf{\xi}_{R} \!\in\! \mathbb{R}^{3\!\times\!3}$ and $\mathbf{\xi}_{T} \!\in\! \mathbb{R}^{3}$ denote the object rotation matrix and translation vector, respectively. 
The superscripts $0$ and $t$ indicate the frame index.
The object is labeled as grasped if the computed errors exceed a defined threshold.

\vspace{-4mm}
\paragraph{Object Switcher.}
With the prepared labels, we employ a multi-layer perceptron (MLP) $g(\cdot)$ to predict the grasping status from the object feature $\mathbf{F}^o_{p}$, which is extracted by the Feature Pyramid Network (FPN)~\cite{lin2017feature} from the input image. 
This process is supervised by the binary cross-entropy loss:
\begin{equation}
\small
    \mathcal{L}^{s}\!=\!- \sum_{j=0}^{1} \mathds{1} (\hat{G} = j) \cdot \operatorname{log} \frac{\operatorname{exp}(g(\mathbf{F}^o_{p})_j)}{\sum_{k=0}^{1} \operatorname{exp}({g(\mathbf{F}^o_{p})_k)}}, 
\end{equation}
where $\hat{G}$ is the ground-truth grasping label and $\mathds{1}(\cdot)$ is the indicator function. 
During testing, the object pose estimation branch is deactivated if predicted as non-grasping, providing more accurate responses for hand-object interactions.

\paragraph{Grasp-aware Feature Fusion.}
Previous studies have shown that feature interaction between hand and object can effectively enhance performance in hand-object scenes~\cite{liu2021semi, tse2022collaborative, lin2023harmonious}; however, such interaction can disrupt hand feature learning in the hand-only scene due to the absence of objects (see \cref{tab:validation}).
To mitigate interference from irrelevant object features, we design the grasp-aware feature fusion.
During training, the object feature $\mathbf{F}^o_{r}$ and the hand feature $\mathbf{F}^h_{r}$ produced by RoIAlign~\cite{he2017mask} are concatenated to form the feature $\mathbf{F}^H$ only when the object is predicted as grasped. 
Next, $\mathbf{F}^H$ is processed through a multi-head attention block~\cite{vaswani2017attention}, resulting in the fused hand-object feature $\mathbf{F}^{OH}$:
\begin{equation}
\small
\begin{aligned}
    \mathbf{F}^H &= \operatorname{Concat} (\mathbf{F}^h_{r}, s \mathbf{F}^o_{r} + (1-s) \mathbf{F}^h_{r}) \quad \text{and}\\
   & \mathbf{F}^{OH} = \operatorname{Softmax} (\frac{\mathbf{F}^H {\mathbf{F}^H}^T}{\sqrt{d^H}})\mathbf{F}^H,
\end{aligned}
\end{equation}
where $s\!\!=\!\!\operatorname{Argmax} (g(\mathbf{F}^o_{p}))$ indicates the predicted grasping status. 
$\operatorname{Concat}(\cdot)$ and $\operatorname{Softmax} (\cdot)$ represent the concatenation and soft-max operations along the channel dimension, respectively.
$d^H$ is the channel dimension of $\mathbf{F}^H$.
This approach enables the network to flexibly toggle object outputs while maintaining robust feature representations for the hand across various scenes.
Then, similar to~\cite{lin2023harmonious}, $\mathbf{F}^{OH}$ is fed into an hourglass-structured hand encoder to produce MANO-related features and regress 2D hand joint coordinates.

\subsection{Occlusion-invariant Feature Learning}
\label{sec3.3:feature_enhancement}
\begin{figure*}
    \centering
    \begin{minipage}{.48\textwidth}
        \centering
        \includegraphics[width=\linewidth]{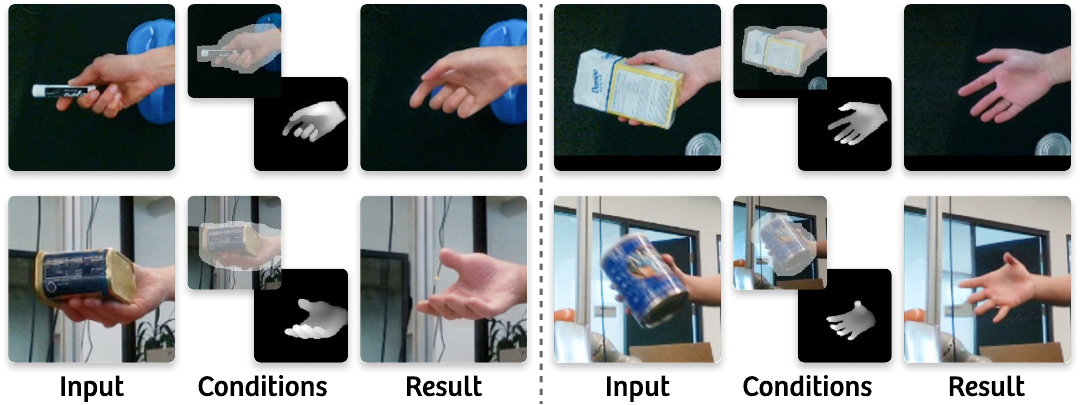}
        % \vspace{-3mm}
        \caption{De-occluded examples in various occlusion conditions.}
        \label{fig:generate_images}
    \end{minipage}%
    \hfill
    \begin{minipage}{.48\textwidth}
        \centering
        \includegraphics[width=\linewidth]{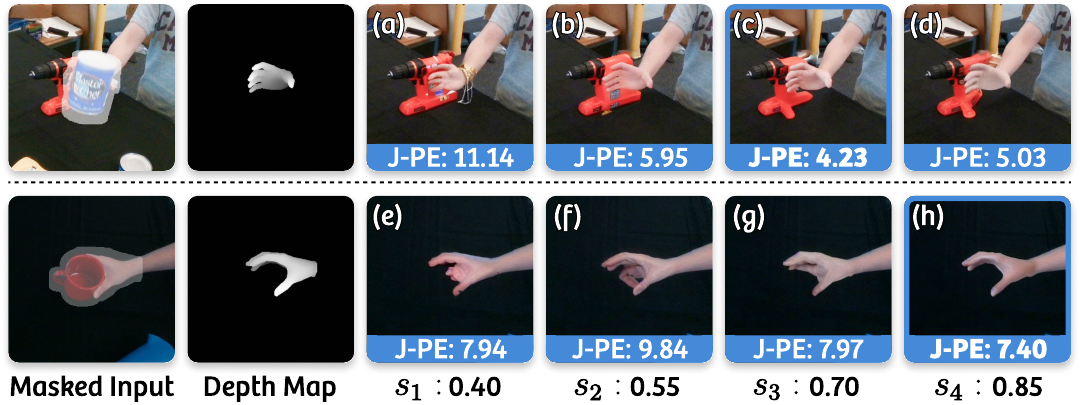}
        % \vspace{-3mm}
        \caption{Visualization of our adaptive control strength adjustment. 
        }
        \label{fig:control_strengths}
    \end{minipage}
    % \vspace{-3mm}
\end{figure*}

Hands are frequently occluded when interacting with objects. 
To achieve robust estimation, the extracted hand feature for the same hand pose should be occlusion-invariant and irrespective of object presence. 
Given the fact that estimating the bare hand pose is easier than that of the one occluded by a held object,
our key insight is to enable the network to simulate the non-occluded hand features from the occluded ones by transferring cross-domain knowledge.
Thus, we propose generating plausible de-occluded hand images as pairs for training samples affected by object-caused occlusions, employing an adaptive adjustment strategy for control strength to maximize generation quality.
Further, we design multi-level feature enhancement techniques that leverage this paired data to promote comprehensive hand feature learning.
% 
% \vspace{-4mm}
\paragraph{Generative De-occluder.}
For occluded hand images, our goal is to de-occlude by realistically removing the grasped object while preserving the hand pose to create paired data. 
Inspired by~\cite{lu2024handrefiner, lugmayr2022repaint}, we utilize ControlNet~\cite{zhang2023adding}, pre-trained on synthetic hand images, to repaint the hand-object region $\mathbf{M}$ guided by a rendered hand depth map $\mathbf{D}$. 
Specifically, following the latent diffusion model~\cite{rombach2022high}, the original image $\mathbf{X}$ is first projected into latent space as $\mathbf{x}_0$ using a variational auto-encoder~\cite{kingma2013auto}. 
We then follow the standard forward diffusion process as outlined in~\cite{ho2020denoising}.
In each reverse step $t\in\{T, T-1, ..., 1\}$, to preserve the known background region $(1-\mathbf{M})\odot\mathbf{X}$, we can alternate the corresponding feature using $(1-\mathbf{m})\odot\mathbf{x}_t$ as long as maintaining the correct properties of its distribution, as the transition from $\mathbf{x}_t$ to $\mathbf{x}_{t-1}$ depends solely on $\mathbf{x}_t$, \ie,
\begin{equation}
\small
    \mathbf{x}^{bg}_{t-1} \sim \mathcal{N}(\sqrt{\bar{\alpha}_{t}}\mathbf{x}_0, (1-\bar{\alpha}_{t}\mathbf{I})),
\end{equation}
where $\mathbf{m}$ is downsampled from $\mathbf{M}$ for calculations in latent space, and $\odot$ is the element-wise product. 
$\mathcal{N}(\cdot)$ denotes the Gaussian distribution.
$\bar{\alpha}_{t}$ denotes the total noise variance at step $t$, as defined in~\cite{song2020denoising}. 
For the unknown hand-object region $\mathbf{M}\odot\mathbf{X}$, we perform the reverse diffusion process using the DDIM sampler~\cite{song2020denoising}, \ie,
\begin{equation}
\small
    \mathbf{x}^{ho}_{t-1} = \operatorname{DDIM}(\epsilon_{\theta}(\mathbf{x}_t, \mathbf{x}_{mask}, \mathbf{D})),
\end{equation}
where $\mathbf{x}_{mask}$ is the latent feature masked by $\mathbf{m}$. 
$\epsilon_{\theta}(\cdot)$ denotes the denoising model.
Thus, the final expression of $\mathbf{x}_{t-1}$ during one reverse step is:
\begin{equation}
\label{eq:mask}
\small
    \mathbf{x}_{t-1} = \mathbf{m} \odot \mathbf{x}^{ho}_{t-1} + (1-\mathbf{m}) \odot \mathbf{x}_{t-1}^{bg},
\end{equation}
which means $\mathbf{x}_{t-1}^{bg}$ is sampled using the known background pixels, while $\mathbf{x}^{ho}_{t-1}$ is sampled from the bare-hand data distribution. 
They are combined into the new $\mathbf{x}_{t-1}$ using the hand-object mask, ensuring both consistency and realism.
After the iterative reverse process, the final denoised vector $\mathbf{x}_0$ is sent to the decoder~\cite{kingma2013auto} to recover images from latent features.
We show examples of different occlusion conditions along with their de-occluded counterparts in~\cref{fig:generate_images}.

\paragraph{Adaptive Control Strength Adjustment.}
To balance consistency with the condition and realism of the generated hand images, the user often needs to manually adjust the control strength parameter in the generative model. 
We visualize some examples generated using different control strengths in~\cref{fig:control_strengths}. 
In certain cases, such as the top row, too-small control strengths make the generated hand not align well with the depth condition (see (a-b)), while overly-large strengths result in unrealistic appearances (see (d)). 
Conversely, in other cases like the bottom row, a large control strength is needed to ensure correct hand anatomy (see (h)). 
Therefore, a fixed control strength cannot be universally applied, while manually setting the value for each case is impractical.
% 
% Inspired by the adaptive strategy in ~\cite{lu2024handrefiner}, 
To this end, we propose adaptively and automatically adjusting the control strength to enhance generation quality. 
Specifically, we first define candidate control strengths \{$s_1, s_2, ..., s_n \mid 0< s_i\leq 1, \forall i$\} and generate a de-occluded image for each.
% $n$ corresponding images.
% 
Then, we employ a pre-trained hand reconstruction model from ~\cite{chen2022mobrecon} to estimate the 3D hand poses from the generated images and evaluate the J-PE against the ground truth. 
The generated sample with the lowest J-PE is incorporated into the training process with the same ground-truth labels as the original sample, \eg, (c) and (h) in ~\cref{fig:control_strengths} are selected.
% based on the lowest J-PE.
% 
This approach maximizes the generation quality by selecting a proper control strength that best balances realism and consistency for each case.

\paragraph{Multi-level Feature Enhancement.}
Given the pair of the original image and the corresponding generated image $(\mathbf{X}, \mathbf{X}^{\prime})$, our goal is to enhance the hand feature representation of $\mathbf{X}$ by leveraging information from $\mathbf{X}^{\prime}$. 
To achieve this, we introduce pair-wise feature constraints within a single network in a self-distillation manner. 
To holistically enhance the capability of recovering occluded information for hand, as shown in \cref{fig:pipeline}, we enhance hand features throughout the hand branch at multiple levels: 
(i) The initial-level feature $\mathbf{F}_{i}^h$, extracted from the initial layer of the FPN, captures the low-level information of the hand; 
(ii) the RoI-level feature $\mathbf{F}_{r}^h$, output from the FPN after the RoIAlign operation, is utilized for adaptive fusion with the object feature; and
(iii) the MANO-level feature $\mathbf{F}_{m}^h$, extracted before the MANO decoder, serves as the most pertinent feature for regressing the MANO coefficients. 
Since the MANO-level feature is at a relatively late stage, the occluded and de-occluded counterparts may not always reside in a similar feature space. 
Inspired by~\cite{chen2017learning}, we adopt a multi-head attention block $h(\cdot)$ as the adaptation layer to improve knowledge transfer. 
Overall, the feature enhancement constraints are formulated as the L1 loss between the features of $\mathbf{X}$ and $\mathbf{X}^{\prime}$:
\begin{equation}
\small
\begin{aligned}
    & \mathcal{L}_{init}^{enh} = ||\mathbf{F}_{i}^h-{\mathbf{F}_{i}^h}^{\prime}||_1,\quad
    \mathcal{L}_{\!\scriptscriptstyle{RoI}}^{enh} = ||\mathbf{F}_{r}^h-{\mathbf{F}_{r}^h}^{\prime}||_1,\\
    & \text{and}\quad \mathcal{L}_{\!\scriptscriptstyle{M\!A\!N\!O}}^{enh} = ||h(\mathbf{F}_{m}^h)-h({\mathbf{F}_{m}^h}^{\prime})||_1,
\end{aligned}
\end{equation}
where features with primes belong to the generated image.
\vspace{-4mm}
\paragraph{Occlusion-aware Case Filtering.}
During the knowledge transition from the generated samples to the original ones, we observe that the feature learning process may not benefit if the original hand is already non-occluded. 
In this case, the knowledge gap between de-occluded and occluded hand features disappears, which causes the feature constraints to focus on mitigating the sim-to-real domain gap, making the original features close to the simulated ones, thus yielding suboptimal performance. 
To address this issue, we filter out non-occluded samples for paired feature enhancement. 
Specifically, we compute the Intersection over Union (IoU) of the provided amodal hand mask (considering object-caused occlusions) and the rendered full hand mask as the ground-truth occlusion proportion. 
During training, feature constraints are exclusively applied to pairs whose original occlusion proportion $\hat{O}$ exceeds a pre-defined threshold $\tau$. 
Furthermore, non-grasping samples are excluded as they do not contribute additional information. 
Thus, the feature enhancement is conducted only on grasping and occluded samples, \ie,
\begin{equation}
    \small
    \mathcal{L}_{*}^{enh} = \mathds{1} ((\hat{O} \geq \tau) \land (\hat{G} = 1)) \cdot \mathcal{L}_{*}^{enh},
\end{equation}
where $\tau\!=\!0.1$ in our experiments. $\mathcal{L}_{*}^{enh}$ represents the feature enhancement constraints at multiple levels.
\begin{table*}[!t]
    \centering
    \resizebox{\linewidth}{!}{
        \begin{tabular}{r@{\hskip 6pt}|l@{\hskip 6pt}|cccc|cccc|cccc}
        \toprule
        &\multicolumn{1}{c|}{\multirow{2}{*}{Methods}} & \multicolumn{4}{c|}{\textit{All Scenes}} & \multicolumn{4}{c|}{\textit{Hand-Only Scene}} & \multicolumn{4}{c}{\textit{Hand-Object Scene}} \\
        \cmidrule{3-14}
          & & J-PE $\downarrow$ & PA-J-PE $\downarrow$  & V-PE $\downarrow$ & PA-V-PE $\downarrow$ & J-PE $\downarrow$ & PA-J-PE $\downarrow$  & V-PE $\downarrow$ & PA-V-PE $\downarrow$ & J-PE $\downarrow$ & PA-J-PE $\downarrow$ & V-PE $\downarrow$ & PA-V-PE $\downarrow$ \\
        \midrule
        &HandOccNet~\cite{park2022handoccnet} & \underline{14.02} & 6.17 & \underline{13.55} & 5.95 & \underline{13.16} & 5.31 & \underline{12.70} & 5.11  & \underline{14.58} & 6.73 & \underline{14.10} & 6.49 \\
        &MobRecon~\cite{chen2022mobrecon}  & 15.02 & 6.71 & 13.93 & 5.91   &  14.43 & 5.88 & 13.45 & 5.15 & 15.40 & 7.25 & 14.24 & 6.39 \\
        &H2ONet~\cite{xu2023h2onet}  & 14.78 & \underline{5.72} & 14.63 & 6.19  &  14.14   & \underline{4.74}  & 14.00 & 5.35 & 15.20 & \underline{6.35} & 15.03 & 6.74 \\
        \multirow{-4}{*}{\rotatebox{90}{HPE}} &SimpleHand~\cite{zhou2024simple}  & 14.78 & 6.30 & 14.11 & 6.03  &  14.63   & 5.62  & 13.96 & 5.38 & 14.88 & 6.74 & 14.21 & 6.45 \\
        \midrule
        &Liu~\etal~\cite{liu2021semi}   & 15.33  & 6.17 & 14.79 & 5.98 & 15.18 & 5.48 & 14.60 & 5.31 & 15.43 & 6.61 & 14.91 &  6.40  \\
        &Keypoint Trans.~\cite{hampali2022keypoint}  & 19.16 & 7.70 & 18.71  &  7.96   & 19.75  & 7.59 & 19.26 & 7.98 & 18.79 & 7.77 & 18.35 & 7.94 \\
        \multirow{-3}{*}{\rotatebox{90}{HOPE}} &HFL-Net~\cite{lin2023harmonious} & 14.32 & 6.08 & 13.83 & 5.86 &  13.61   &  5.20 & 13.10 & \underline{5.01} & 14.77 & 6.64 & 14.29 & 6.41\\
        \midrule
        &H2ONet$^\dagger$ + HFL-Net$^\dagger$   & 14.12  & 5.83 & 13.75 & \underline{5.83} & 13.29 & \textbf{4.70} & 13.08 & 5.06  & 14.66 & 6.55 & 14.18 & \underline{6.33}   \\
        &H2ONet$^\ddagger$ + HFL-Net$^\ddagger$   & 14.54  & 5.90 & 14.19 & 6.00 & 14.14 & 4.76 & 14.00 & 5.35  & 14.79 & 6.63 & 14.31 & 6.41   \\
        &HandOccNet$^\dagger$ + HFL-Net$^\dagger$   & 14.34  & 6.06 & 13.86 & 5.85 & 13.85 & 5.28 & 13.35 & 5.08  & 14.66 & 6.56 & 14.18 & 6.34   \\
        &HandOccNet$^\ddagger$ + HFL-Net$^\ddagger$   & 14.52  & 6.15 & 14.02 & 5.93 & 14.09 & 5.37 & 13.58 & 5.17  & 14.79 & 6.65 & 14.31 & 6.42   \\
        \multirow{-5}{*}{\rotatebox{90}{Unified}} & UniHOPE (ours) & \textbf{13.03} & \textbf{5.59} & \textbf{12.59} & \textbf{5.40} &  \textbf{12.59}   & 4.83  & \textbf{12.12} & \textbf{4.66} & \textbf{13.31} & \textbf{6.08} & \textbf{12.89} & \textbf{5.87} \\
        \bottomrule
        \end{tabular}
        }
    \vspace{-3mm}
    \caption{Hand-pose estimation results on DexYCB.
    $^\dagger$: pre-trained in the original setting.
    $^\ddagger$: re-trained in the unified setting.
    The best and second-best are marked in \textbf{bold} and \underline{underlined}.
    Our UniHOPE attains leading performance for almost all metrics in all scenarios.
    }
    \label{tab:dexycb_s3}
    \vspace{-3mm}
\end{table*}

\subsection{Loss Functions}
\label{sec3.4:loss}
Our training loss includes regular hand and object losses similar to~\cite{lin2023harmonious}, along with our object switcher loss $\mathcal{L}^s$, and feature enhancement constraints $\mathcal{L}_{init}^{enh}$, $\mathcal{L}_{\!\scriptscriptstyle{RoI}}^{enh}$, and $\mathcal{L}_{\!\scriptscriptstyle{M\!A\!N\!O}}^{enh}$.
First, the hand loss is computed as $\mathcal{L}^{h} \!=\! \mathcal{L}^{J} \!+\! \mathcal{L}^{V} \!+\! \mathcal{L}^{\scriptscriptstyle{M\!A\!N\!O}}$, where 
\begin{equation}
    \small
    \begin{aligned}
    & \mathcal{L}^{J} = ||\mathbf{J}^{2D} \!-\! \mathbf{\hat{J}}^{2D}||_{2} + ||\mathbf{J}^{3D} \!-\! \mathbf{\hat{J}}^{3D}||_{2}, \\ 
     \mathcal{L}^{V} =& ||\mathbf{V} \!-\! \mathbf{\hat{V}}||_{2}, \ \ \text{and} \ \ \mathcal{L}^{\scriptscriptstyle{M\!A\!N\!O}} = ||(\theta; \beta) \!-\! (\hat{\theta}; \hat{\beta})||_{2}. \\
    \end{aligned}
\end{equation}
Here $\mathbf{J}^{2D}$, $\mathbf{J}^{3D}$, and $\mathbf{V}$ represent the 2D joint, 3D joint, and 3D vertex coordinates, respectively.
$(\theta; \beta)$ represent the MANO coefficients.
The hat superscript denotes the ground-truth label.

The object loss $\mathcal{L}^{o}$ supervises the predictions of the 2D location (projected from 3D object keypoints) and their corresponding confidences from image grid proposals~\cite{redmon2016you}, \ie,
\begin{equation}
\small
    \mathcal{L}^{o} = \sum_{g}\sum_{k=1}^{N^o} (||p_{g,k} - \hat{p}_{g,k}||_1 + ||c_{g,k} - \hat{c}_{g,k}||_1),
\end{equation}
where $N^o$ is the number of keypoints in the 3D bounding box of object mesh. $p_{g,k}$ and $c_{g,k}$ are the pixel location and confidence value at the grid $g$ and control point $k$, respectively. The hat superscript denotes the ground truth. 
We compute the object loss only for those grasping images, as they contain the complete object for pose estimation.

Overall, the training loss is as follows,
\begin{equation}
\small
\begin{aligned}
    \mathcal{L}^{total} = \;& \begin{matrix} \underbrace{\mathcal{L}^{h} + \mathcal{L}^{o} + \alpha \mathcal{L}^{s}} \\ \text{sample-wise}\end{matrix} \\
    + \;& \begin{matrix} \underbrace{\gamma_{init} \mathcal{L}_{init}^{enh} + \gamma_{\scriptscriptstyle{RoI}} \mathcal{L}_{\!\scriptscriptstyle{RoI}}^{enh} + \gamma_{\scriptscriptstyle{M\!A\!N\!O}} \mathcal{L}_{\!\scriptscriptstyle{M\!A\!N\!O}}^{enh}} \\ \text{pair-wise} \end{matrix}
\end{aligned}
\label{eq:loss_functions}
\end{equation}
where $\alpha$ and $\gamma_{*}$ are weights to balance the loss terms.
\begin{table}[t]
    \centering
    \resizebox{\linewidth}{!}{
        \setlength\tabcolsep{2.2pt}
        \begin{tabular}{l|cccccc}
        \toprule
         \multicolumn{1}{c|}{Methods} & gelatin\_box & bleach\_cleanser    & wood\_block & average $\uparrow$  \\
        \midrule
        Liu~\etal~\cite{liu2021semi}  & 26.31 & 25.07 & 68.56 & 38.89  \\
        Keypoint Trans.~\cite{hampali2022keypoint}  & 0.00 & 1.31 & 32.61 & 10.47  \\
        HFL-Net~\cite{lin2023harmonious} & 25.88 & \underline{32.08} & \underline{70.16} & \underline{41.59}  \\
        H2ONet$^\dagger$ + HFL-Net$^\dagger$  & \underline{29.26} & 30.71 & 64.84 & 40.69  \\
        H2ONet$^\ddagger$ + HFL-Net$^\ddagger$  & 26.12 & 29.33 & 69.40 & 40.51  \\
        HandOccNet$^\dagger$ + HFL-Net$^\dagger$  & \textbf{29.30} & 30.57 & 64.94 & 40.69  \\
        HandOccNet$^\ddagger$ + HFL-Net$^\ddagger$  & 26.22 & 29.51 & 69.40 & 40.61  \\
        \ourname{} (ours) & 26.23 & \textbf{32.32} & \textbf{74.29} & \textbf{43.06}  \\
        \bottomrule
        \end{tabular}}
    \vspace{-3mm}
    \caption{Unseen object-pose estimation results on DexYCB.}
    \label{tab:dexycb_s3_object}
    \vspace{-3mm}
\end{table}

\begin{table}[!t]
    \resizebox{\linewidth}{!}{
        \setlength\tabcolsep{2pt}
        \begin{tabular}{r@{\hskip 6pt}|l@{\hskip 6pt}|cccccc}
        \toprule
        \cmidrule{3-7}
          &  \multicolumn{1}{c|}{Methods} & J-PE $\downarrow$ & J-AUC $\uparrow$ & V-PE $\downarrow$ & V-AUC $\uparrow$ & F@5 $\uparrow$ & F@15 $\uparrow$ \\
        \midrule
        &HandOccNet~\cite{park2022handoccnet} & 28.94 & 47.96 & 28.10 & 49.23 & 23.34 & 67.86  \\
        &MobRecon~\cite{chen2022mobrecon}  & 29.61 & 48.10 & 28.65 & 49.59  & 23.38 & 67.72 \\
        &H2ONet~\cite{xu2023h2onet}  & 30.46 & 47.09 & 29.55 & 48.31  & 22.10  & 66.04 \\
        \multirow{-4}{*}{\rotatebox{90}{HPE}} &SimpleHand~\cite{zhou2024simple}  & 29.01 & 47.58 & 28.09 &  49.00 &  21.93   & 65.90 \\
        \midrule
        &Liu~\etal~\cite{liu2021semi}   & 29.54 & 47.55 & 28.66 & 48.83  & 21.82  & 67.34  \\
        &Keypoint Trans.~\cite{hampali2022keypoint}  & 41.04 & 34.47 & 39.64 & 36.02  &  17.13  & 56.07 \\
        \multirow{-3}{*}{\rotatebox{90}{HOPE}} &HFL-Net~\cite{lin2023harmonious} & 28.45 & 50.34 & 27.55 & 51.57 &  24.31  & 69.88 \\
        \midrule
        &H2ONet$^\dagger$ + HFL-Net$^\dagger$   & 31.27 & 47.70 & 30.31 & 48.86  & 22.91  & 67.20 \\
        &H2ONet$^\ddagger$ + HFL-Net$^\ddagger$  & 28.49 & \underline{50.45} & 27.59 & \underline{51.67}  &  24.33   & 69.86 \\
        &HandOccNet$^\dagger$ + HFL-Net$^\dagger$  & 30.96 & 47.85 & 30.01 & 49.02  &  23.04   & 67.47 \\
        &HandOccNet$^\ddagger$ + HFL-Net$^\ddagger$  & \underline{28.33} & \underline{50.45} & \underline{27.44} & \underline{51.67}  &  \underline{24.39}   & \underline{69.99} \\
        \multirow{-5}{*}{\rotatebox{90}{Unified}} &\ourname{} (ours) & \textbf{26.23} & \textbf{52.26} & \textbf{25.41} & \textbf{53.52} &  \textbf{24.64}   & \textbf{70.77} \\
        \bottomrule
        \end{tabular}
    }
    \vspace{-3mm}
    \caption{Hand-pose estimation results (\emph{Root-relative}) on HO3D.}
    \label{tab:ho3d_v2}
    \vspace{-3mm}
\end{table}

\section{Experiments}
\label{sec4:exp}
\subsection{Experimental Settings}
\paragraph{Datasets.}
In our unified setting, we organize the original dataset into hand-only and hand-object scenes based on the object grasping status.
We conduct experiments on the following commonly-used datasets:
(i) \textbf{DexYCB}~\cite{chao2021dexycb}: we use the more challenging ``S3'' split (train/test: 376,374/76,360 samples) with unseen grasped objects in the test set (train/test: 15/3 objects). 
We report performance on the entire dataset (all scenes) as well as separately for hand-only and hand-object scenes; 
(ii) \textbf{HO3D}~\cite{hampali2020honnotate} (version 2, train/test: 66,034/11,524 samples):
results are submitted to the online server as the ground-truth 3D hand annotations are not publicly accessible, hence results for each scene are unavailable. 
Further, to evaluate the generalization ability, we perform cross-dataset validation on the test set of 
(iii) \textbf{FreiHAND}~\cite{zimmermann2019freihand} (train/test: 130,240/3,960 samples), which mainly consists of bare-hand images and lacks object annotations for scene division.
Consequently, results for each scene are also unavailable.
More details and results on other data splits of DexYCB are provided in the Supp.
\begin{figure*}[t]
    \centering
    \includegraphics[width=\linewidth]{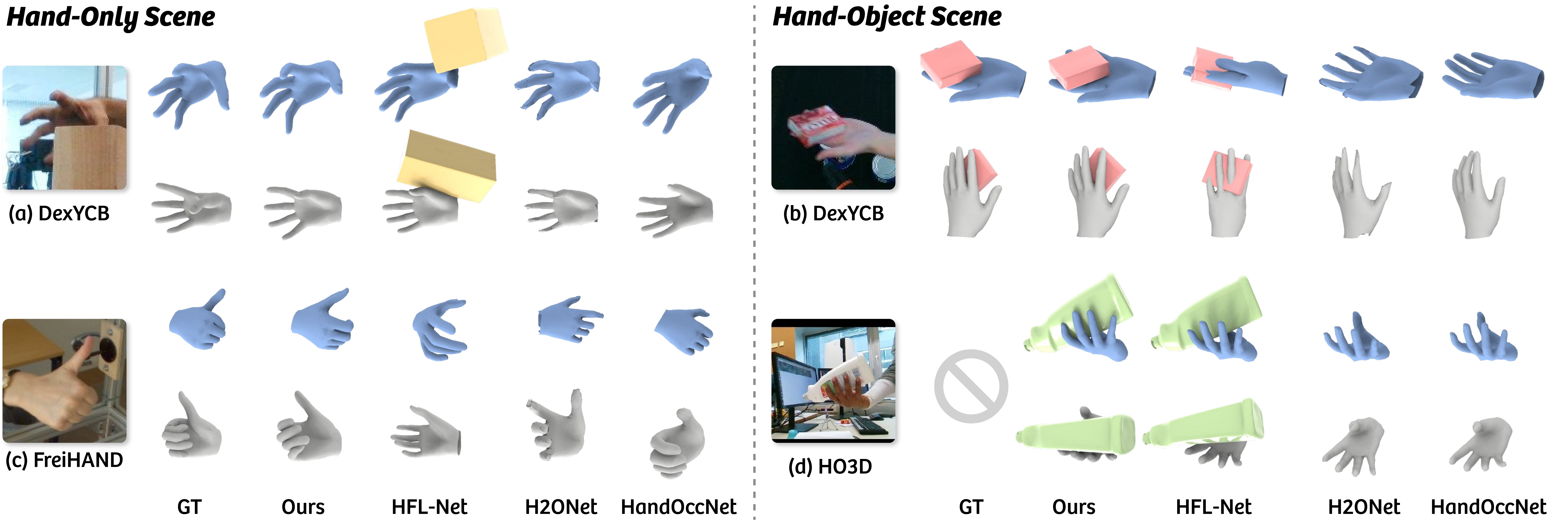}
    \vspace{-7mm}
    \caption{Qualitative comparison between our method and SOTA HPE/HOPE methods on hand-only/hand-object scenarios across different datasets. The first and second rows in each example denote the original view and another view, respectively, for better comparison.}
    \label{fig:qualitative_results}
    \vspace{-3mm}
\end{figure*}
\begin{figure}[t]
    \centering
    \includegraphics[width=\linewidth]
    {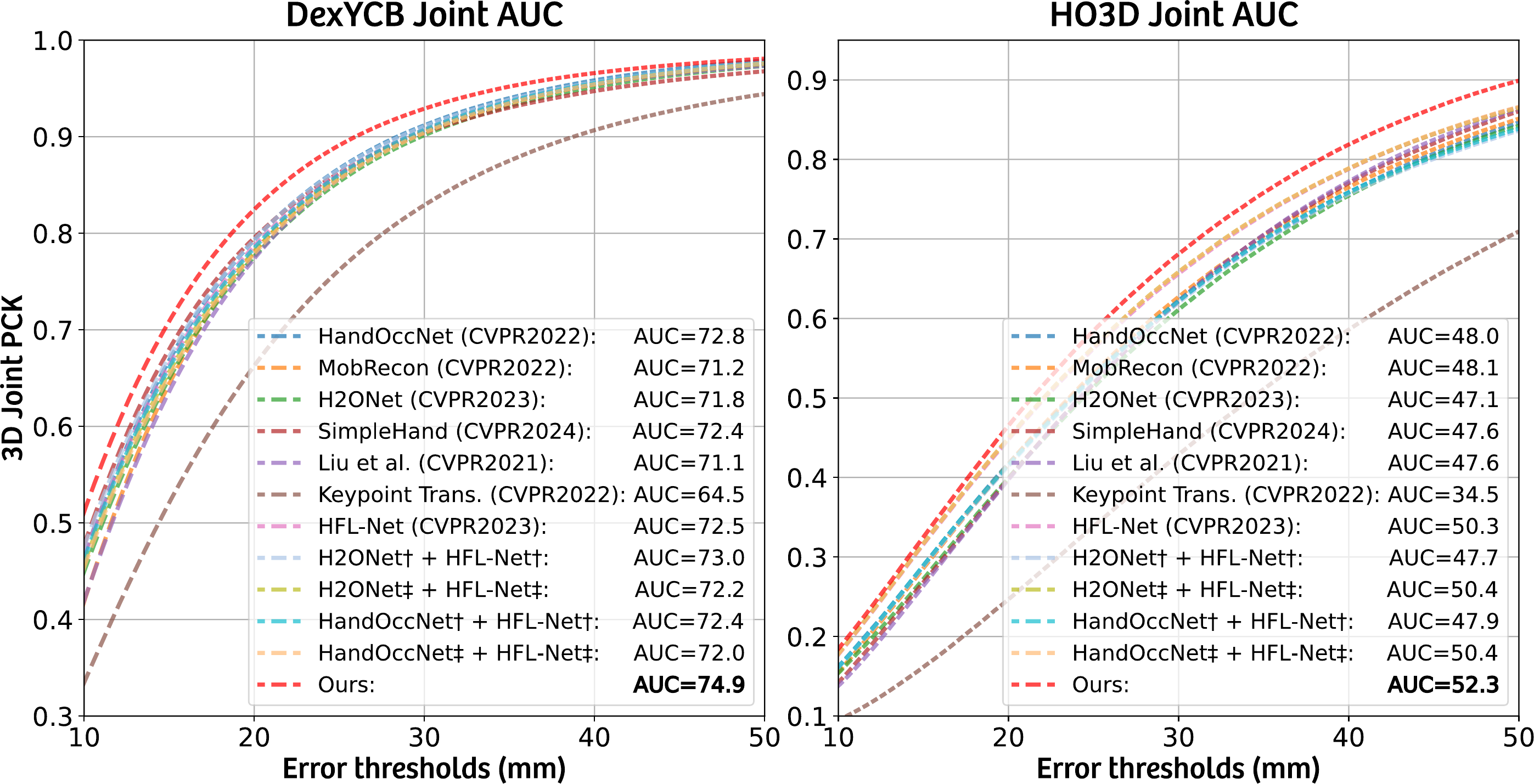}\\
    \vspace{-3mm}
    \caption{The joint AUC comparison under different thresholds. 
    \ourname{} achieves better performance than others, consistently.
    }
    \label{fig:pck}
    \vspace{-3mm}
\end{figure}

\vspace{-4mm}
\paragraph{Evaluation Metrics.}
We evaluate hand pose estimation using commonly-used metrics, as in~\cite{park2022handoccnet, chen2022mobrecon, xu2023h2onet}:
(i) J/V-PE denotes the mean per joint/vertex position error (also known as MPJPE/MPVPE) in mm measured by Euclidean distance between estimated and ground-truth 3D hand joint/vertex coordinates; 
(ii) J/V-AUC calculates the area under the curve (in percentage) of the percentage of correct keypoints (PCK) across different error thresholds for joint/vertex; and
(iii) F@5/F@15 is the harmonic mean of recall and precision (in percentage) between estimated and ground-truth 3D hand vertices under 5mm/15mm thresholds. 
% 
% Besides, 
We report these metrics both before (\ie, root-relative) and after Procrustes Alignment (PA), which aligns the estimation with ground truths by global rotation, translation, and scale adjustment.

For object pose evaluation, we measure the average 3D distance (ADD) of the grasped object.
% % 
%
% For DexYCB, 
Specifically, we report ADD-0.5D, the percentage of objects whose ADD is within 50\% of the object diameter as in~\cite{hu2022perspective} considering the challenge of unseen-object pose estimation in DexYCB.
\vspace{-4mm}
\paragraph{Implementation Details.}
We train \ourname{} on eight NVidia RTX 2080Ti GPUs using a batch size of 64 and the Adam optimizer~\cite{kingma2014adam} with an initial learning rate of 1e-4 (decay by 0.7 every 10 epochs). 
Input images are resized to $128\!\times\!128$ and augmented with random scaling, rotating, translating, and color jittering. 
To stabilize training, we first train the network with both original and generated images for 30 epochs, then incorporate the feature enhancement constraints for another 40 epochs.
Please refer to our Supp. for more details.

\subsection{Comparison with SOTA Methods}
We compare our \ourname{} with previous SOTA HPE and HOPE methods~\cite{park2022handoccnet, chen2022mobrecon, zhou2024simple, liu2021semi, hampali2022keypoint} in our unified setting (trained using their officially-released code).
To support the unified task, one straightforward solution is to use a classifier to determine if the hand is grasping any object, then employ an existing HPE or HOPE method accordingly. 
We denote the combination as A + B, where A is the SOTA HPE method \text{H2ONet}~\cite{xu2023h2onet} or \text{HandOccNet}~\cite{park2022handoccnet}, and B is the SOTA HOPE method, HFL-Net~\cite{lin2023harmonious}. 

\vspace{-4mm}
\paragraph{Evaluation on DexYCB.}
We present quantitative comparisons of hand pose estimation on DexYCB in~\cref{tab:dexycb_s3}. 
Our \ourname{} achieves the best performance overall, showing its effectiveness in general hand-object interactions across various scenarios. 
The root-relative 3D joint PCK/AUC comparison under different thresholds is shown in~\cref{fig:pck}, further confirming the comprehensive performance of our approach.

For object pose estimation accuracy, we conduct comparisons using per-instance and average ADD-0.5D, as shown in~\cref{tab:dexycb_s3_object}.
Compared with SOTA HOPE methods, our method achieves the highest average score on the test set with unseen objects, highlighting its superiority in estimating object pose. 
Qualitative comparisons with SOTA methods~\cite{park2022handoccnet, xu2023h2onet, lin2023harmonious} are illustrated in~\cref{fig:qualitative_results}.
In the hand-only scenario (see \cref{fig:qualitative_results} (a)), where the wood block has not been grasped yet, SOTA HOPE method~\cite{lin2023harmonious} inevitably produces an extra object pose due to its inflexibility; in contrast, our method does not, thanks to our object switcher's >95\% grasping status prediction accuracy. 
Moreover, our method yields more plausible hand poses.
In the hand-object scenario (see \cref{fig:qualitative_results} (b)), our method produces high-quality hand-object poses, whereas previous methods fail when the hand experiences moderate occlusion, indicating our robustness against such challenges. 

\vspace{-4mm}
\paragraph{Evaluation on HO3D.}
We conduct the same experiment on HO3D~\cite{hampali2020honnotate}. 
\cref{tab:ho3d_v2} shows the root-relative quantitative comparison. 
Our method achieves top performance across all metrics, demonstrating its effectiveness and robustness. 
The joint PCK/AUC curve in~\cref{fig:pck} also confirms the consistent best results of our approach.
In addition, qualitative comparisons in~\cref{fig:qualitative_results} (d) clearly illustrate the superiority of our method in estimating 3D hand and object pose under partial object-caused occlusion. 
More quantitative results are available in the Supp.

\vspace{-4mm}
\paragraph{Evaluation on FreiHAND.}
To assess generalization ability, we perform cross-dataset validation by transferring models trained on DexYCB to the FreiHAND test set. 
As reported in~\cref{tab:freihand}, our method outperforms all SOTA HPE/HOPE methods, particularly by a substantial margin in root-relative metrics.
The qualitative comparison in~\cref{fig:qualitative_results} (c) also shows that \ourname{} produces more accurate hand poses in challenging cases, indicating improved generalization to unseen bare-hand scenes. 

\vspace{-4mm}
\paragraph{Evaluation under Different Levels of Occlusion.}
To showcase the robustness of our method against object-caused occlusion, we partition the DexYCB test set into different occlusion levels based on the ground-truth hand-object occlusion proportion (as detailed in \cref{sec3.3:feature_enhancement}) and provide quantitative comparisons in~\cref{tab:occ_level}. 
Our \ourname{} exhibits the best hand pose estimation performance across all occlusion levels, underscoring the efficacy of our feature enhancement techniques. 
% 
% Note that test samples with fully-occluded hands are excluded. 
Note that test samples where the hand being absent from the image region are excluded. 

\begin{table*}[!t]
    \centering
    \resizebox{\linewidth}{!}{
        \begin{tabular}{r@{\hskip 6pt}|l@{\hskip 6pt}|cccccc| cccccc}
        \toprule
        &\multicolumn{1}{c|}{\multirow{2}{*}{Methods}} & \multicolumn{6}{c|}{\textit{Root-relative}} & \multicolumn{6}{c}{\textit{Procrustes Alignment}} \\
        \cmidrule{3-14}
          & & J-PE $\downarrow$ & J-AUC $\uparrow$ & V-PE $\downarrow$ & V-AUC $\uparrow$ & F@5 $\uparrow$ & F@15 $\uparrow$ & J-PE $\downarrow$ & J-AUC $\uparrow$ & V-PE $\downarrow$ & V-AUC $\uparrow$ & F@5 $\uparrow$ & F@15 $\uparrow$ \\
        \midrule
        &HandOccNet~\cite{park2022handoccnet} & 58.03 & 29.07 & \underline{56.06} & 29.40 & 14.89 & 47.75 & 14.52 & 71.34 & 14.09 & 72.07 & 40.02 & 88.08 \\
        &MobRecon~\cite{chen2022mobrecon}  & 71.18 & 22.21 & 68.50 & 22.34  & 10.41  & 36.67 & 17.74 & 65.16 & 17.32 & 65.90 & 32.95 & 81.35 \\
        &H2ONet~\cite{xu2023h2onet}  & 79.14 & 19.88 & 76.29 & 19.38  & 9.69 & 36.71 & 14.56 & 71.21 & 14.25 & 71.69 & 38.73 & 87.69 \\
        \multirow{-4}{*}{\rotatebox{90}{HPE}} &SimpleHand~\cite{zhou2024simple}  & 60.82 & 26.09 & 58.85 & 26.34  &  12.84 & 43.53  & 15.79 & 68.90 & 15.37 & 69.61 & 37.03 & 85.40 \\
        \midrule
        &Liu~\etal~\cite{liu2021semi}   & 59.40 & 28.84 & 57.39 & 29.04  & 14.92  & 47.76 & \underline{14.00} & \underline{72.30} & \underline{13.59} & \underline{73.02} & \underline{41.53} & \underline{89.11} \\
        &Keypoint Trans.~\cite{hampali2022keypoint}  & 96.27 & 13.60 & 93.39 & 12.37  & 6.99 & 27.82 & 16.97 & 66.66 & 17.03 & 66.38 & 34.31 & 83.44 \\
        \multirow{-3}{*}{\rotatebox{90}{HOPE}} &HFL-Net~\cite{lin2023harmonious} & \underline{58.02} & \underline{29.94} & 56.08 & \underline{30.20} &  \underline{15.47} & \underline{48.83} & 14.29 & 71.75 & 13.85 & 72.52 & 40.52 & 88.63 \\
        \midrule
        &H2ONet$^{\dagger}$ + HFL-Net$^\dagger$  & 68.88 & 24.27 & 66.49 & 24.21 & 12.33 & 42.38  & 14.47 & 71.36 & 14.10 & 71.99 & 39.48 & 88.14 \\
        &H2ONet$^{\ddagger}$ + HFL-Net$^\ddagger$  & 68.25 & 24.49 & 65.90 & 24.40 &  12.30  & 42.51 & 14.42 & 71.47 & 14.06 & 72.09 & 39.54 & 88.18 \\
        &HandOccNet$^\dagger$ + HFL-Net$^\dagger$  & 81.79 & 21.33 & 78.91 & 20.56  &  10.89  & 36.68 & 15.40 & 69.65 & 14.85 & 70.58 & 38.30 & 86.94 \\
        &HandOccNet$^\ddagger$ + HFL-Net$^\ddagger$  & 80.74 & 20.30 & 78.00 & 19.44 &  10.52 & 36.38 & 16.15 & 68.29 & 15.58 & 69.26 & 36.92 & 85.93 \\
        \multirow{-3}{*}{\rotatebox{90}{Unified}} &\ourname{} (ours) & \textbf{50.97} & \textbf{34.31} & \textbf{49.21} & \textbf{34.94} & \textbf{17.83} & \textbf{53.46} & \textbf{13.53} & \textbf{73.24} & \textbf{13.14} & \textbf{73.92} & \textbf{43.23} & \textbf{89.55} \\
        \bottomrule
        \end{tabular}
        }
    \vspace{-3mm}
    \caption{Cross-dataset validation of hand-pose estimation on FreiHAND.}
    \label{tab:freihand}
    \vspace{-2mm}
\end{table*}
\begin{table*}[!t]
    \centering
    \resizebox{\linewidth}{!}{
        \setlength\tabcolsep{4pt}
        \begin{tabular}{r@{\hskip 6pt}l@{\hskip 6pt}|cccc|cccc|cccc}
        \toprule
        & \multicolumn{1}{c|}{\multirow{2}{*}{Methods}} & \multicolumn{4}{c|}{\textit{Occlusion (25\%-50\%)}} & \multicolumn{4}{c|}{\textit{Occlusion (50\%-75\%)}} & \multicolumn{4}{c}{\textit{Occlusion (75\%-100\%)}} \\
        \cmidrule{3-14}
          & & J-PE$\downarrow$ & PA-J-PE$\downarrow$ & V-PE $\downarrow$ & PA-V-PE $\downarrow$ & J-PE$\downarrow$ & PA-J-PE$\downarrow$ & V-PE $\downarrow$ & PA-V-PE $\downarrow$ & J-PE$\downarrow$ & PA-J-PE$\downarrow$ & V-PE $\downarrow$ & PA-V-PE $\downarrow$ \\
        \midrule
        &HandOccNet~\cite{park2022handoccnet} & 16.40 & 7.08 & 15.85 & 6.83 & \underline{18.22} & 7.60 & \underline{17.67} & 7.33 & \underline{28.15} & 8.71 & \underline{27.20} & 8.40 \\
        &MobRecon~\cite{chen2022mobrecon}   & 16.67 & 7.61 & 15.60 & 6.77 & 20.04 & 8.17 & 18.80 & 7.48 & 31.46 & 9.64 & 29.97 & 9.23 \\
        &H2ONet~\cite{xu2023h2onet}     &  17.07 & \underline{6.76} & 16.78 & 7.09 & 19.41 & 7.32 & 19.07 & 7.58 & 31.07 & 8.82 & 30.11 & 8.94 \\
        &SimpleHand~\cite{zhou2024simple}     & 16.43 & 7.05 & 15.78 & 6.75 & 19.33 & 7.55 & 18.38 & 7.39 & 38.52 & 10.57 & 36.85 & 10.40 \\
        &Liu~\etal~\cite{liu2021semi}     &  16.98 & 6.92 & 16.43 & 6.71 & 19.72 & \underline{7.11} & 19.14 & \underline{6.91} & 33.80 & 8.99 & 32.64 & 8.71\\
        &Keypoint Trans.~\cite{hampali2022keypoint}     &  20.95 & 8.15 & 20.41 & 8.31 & 24.45 & 8.61 & 23.88 & 8.76 & 38.29 & 11.21 & 37.39 & 11.75\\
        &HFL-Net~\cite{lin2023harmonious}     &  16.33 & 7.00 & 15.81 & 6.77 & 18.66 & 7.33 & 18.11 & 7.11 & 28.95 & 8.80 & 27.94 & 8.53\\
        &H2ONet$^\dagger$ + HFL-Net$^\dagger$     &  16.07 & 6.84 & 15.57 & 6.67 & 19.39 & 7.40 & 18.82 & 7.22 & 30.32 & \underline{8.43} & 29.27 & \underline{8.28} \\
        &H2ONet$^\ddagger$ + HFL-Net$^\ddagger$     &  16.43 & 6.93 & 15.94 & 6.79 & 18.76 & 7.29 & 18.25 & 7.14 & 31.02 & 8.55 & 29.91 & 8.45 \\
        &HandOccNet$^\dagger$ + HFL-Net$^\dagger$     &  \underline{16.04} & 6.89 & \underline{15.53} & \underline{6.65} & 19.22 & 7.43 & 18.64 & 7.20 & 29.57 & 8.58 & 28.53 & 8.31 \\
        &HandOccNet$^\ddagger$ + HFL-Net$^\ddagger$     &  16.41 & 7.00 & 15.88 & 6.77 & 18.64 & 7.33 & 18.09 & 7.10 &  29.22 & 8.76 & 28.20 & 8.47 \\
        & \ourname{} (ours)      &  \textbf{14.59}   & \textbf{6.39}  & \textbf{14.13} & \textbf{6.17} & \textbf{16.27} & \textbf{6.51} & \textbf{15.78} & \textbf{6.29} & \textbf{26.42} & \textbf{7.64} & \textbf{25.51} & \textbf{7.40}\\
        \bottomrule
        \end{tabular}
        }
    \vspace{-3mm}
    \caption{Comparison with SOTA methods across different object-caused occlusion levels on DexYCB.
    }
    \label{tab:occ_level}
    \vspace{-3mm}
\end{table*}

\begin{table}[t]
    \centering
    \resizebox{\linewidth}{!}{%
    \setlength\tabcolsep{2.5pt}
    \begin{tabular}{c|l|cc|cc}
        \toprule
        &\multicolumn{1}{c|}{\multirow{2}{*}{Models}} & \multicolumn{2}{c|}{\emph{Root-relative}} & \multicolumn{2}{c}{\emph{Procrustes Align.}} \\
        \cmidrule{3-6}
        & & J-PE $\downarrow$ & V-PE $\downarrow$ & J-PE $\downarrow$ & V-PE $\downarrow$ \\
        \midrule
        (a) & Baseline & 14.09 & 13.61  & 5.95 & 5.75  \\
        (b) & w/ Grasp-aware Feature Fusion & 13.84 & 13.37 & 5.79 & 5.58 \\
        \midrule
        (c) & w/ Generative De-occluder & 13.38 & 12.92 & 5.71 & 5.52 \\
        \midrule
        (d) & + Image Feature Enhancement & 13.23 & 12.79 & 5.64 & 5.44 \\
        (e) & + RoI Feature Enhancement & 13.18 & 12.73 & 5.64 & 5.45 \\
        (f) & + MANO Feature Enhancement & 13.12 & 12.67 & 5.63 & 5.43 \\
        (g) & + Occlusion-aware Case Filtering & \textbf{13.03} & \textbf{12.59} & \textbf{5.59} & \textbf{5.40} \\
         \bottomrule
    \end{tabular}%
    }
    \vspace{-3mm}
    \caption{Ablation study on major designs in \ourname{}.}
    \label{tab:ablation_study}
    \vspace{-3mm}
\end{table}

\subsection{Ablation Studies}
We perform ablation studies on DexYCB to evaluate the effectiveness of our designs, as shown in~\cref{tab:ablation_study}. 
HFL-Net~\cite{lin2023harmonious} with our object switcher serves as the simplest baseline.

\vspace{-3mm}
\paragraph{Grasp-aware Feature Fusion.} 
We first analyze the impact of the grasp-aware feature fusion module. 
Comparison of Rows (a-b) shows performance boosts across all metrics, indicating that integrating irrelevant object features affects hand pose estimation and our design alleviates this issue.

\vspace{-3mm}
\paragraph{Generative De-occluder.}
Next, we assess the effects of de-occluded hand images with in-distribution hand poses, as they provide extra information.
Comparing Rows (b-c), the notable improvement upon incorporating paired data indicates the effectiveness of synthetic samples. 
More details on our adaptive control strength adjustment are in the Supp.
\vspace{-3mm}
\paragraph{Occlusion-invariant Feature Learning.}
Further, we show the individual effects of feature enhancement at various levels. 
Comparing Row (c) with (d-f), we note a progressive improvement in root-relative metrics at each level, showing the efficacy of knowledge transferring from de-occluded hands.
Finally, Row (g) reveals that the effects are maximally realized through our occlusion-aware case filtering.

\section{Conclusion}
\label{sec5:conclusion}
We introduce~\ourname{}, the first unified approach for hand-only and hand-object pose estimation, 
motivated by the inability of existing methods to handle both scenes. 
Our technical innovations are twofold:
first, to enable flexibility in switching between different scenes, we incorporate an object switcher to control object-pose estimation and design a grasping-aware feature fusion module to selectively capture effective object features; 
second, to promote robustness against object-caused occlusion, we propose multi-level feature enhancement to learn occlusion-invariant hand features from generated realistic de-occluded hand images. Experimental results on three common benchmarks manifest the SOTA performance of~\ourname{}.

\paragraph{Acknowledgments}
This work is supported in part by the Research Grants Council of the Hong Kong Special Administrative Region, China, under Project T45-401/22-N and 
by the Hong Kong Innovation and Technology Fund, under Project MHP/086/21.
% \clearpage
{
    \small
    \bibliographystyle{ieeenat_fullname}
    \bibliography{review}
}


\end{document}

\clearpage
\setcounter{page}{1}
\maketitlesupplementary

\renewcommand{\thetable}{\Alph{table}}
\renewcommand{\thefigure}{\Alph{figure}}
\renewcommand{\thesection}{\Alph{section}}
\newcolumntype{Y}{>{\centering\arraybackslash}X}

% \section{Overview}
In this supplementary material, we provide more qualitative and quantitative results to show the capabilities and robustness of~\ourname{} (\cref{sec:supp_experimental_results}). In~\cref{sec:supp_implementation_details}, we present the implementation details and in~\cref{sec:supp_limitations}, we discuss the limitations and future work.

\section{More Experimental Results}
\label{sec:supp_experimental_results}

\subsection{Qualitative Results}
\label{sec:supp_qualitative_results}
\begin{figure*}[ht]
    \centering
    \includegraphics[width=\linewidth]{figures/img/supp_gallery_dexycb_1.pdf}
    \caption{\ourname{} is able to handle both hand-only (left column) and hand-object scenarios (right column). Here, we show more qualitative results on DexYCB. For each example, the estimation results are rendered from the original (view 1) and another view (view 2) for clear visualization.}
    \label{fig:supp_gallery_dexycb_1}
\end{figure*}

\begin{figure*}[ht]
    % \ContinuedFloat
    \centering
    \includegraphics[width=\linewidth]{figures/img/supp_gallery_dexycb_2.pdf}
    \caption{More qualitative results of~\ourname{} on DexYCB.}
    \label{fig:supp_gallery_dexycb_2}
\end{figure*}
\begin{figure*}[ht]
    \centering
    \includegraphics[width=\linewidth]{figures/img/supp_gallery_ho3d_1.pdf}
    \caption{More qualitative results of~\ourname{} across hand-only (left column) and hand-object scenarios (right column) on HO3D.}
    \label{fig:supp_gallery_ho3d_1}
\end{figure*}

\begin{figure*}[ht]
    % \ContinuedFloat
    \centering
    \includegraphics[width=\linewidth]{figures/img/supp_gallery_ho3d_2.pdf}
    \caption{More qualitative results of~\ourname{} on HO3D.}
    \label{fig:supp_gallery_ho3d_2}
\end{figure*}
First of all, we present~\cref{fig:supp_gallery_dexycb_1,fig:supp_gallery_dexycb_2,fig:supp_gallery_ho3d_1,fig:supp_gallery_ho3d_2}, which show that~\ourname{} is able to handle both hand-only scenario (left columns) and hand-object scenario (right columns).

% \phil{TODO: keep Figures A to D on P.2-5 of this supp.}

\vspace*{-3mm}
\paragraph{Comparison with SOTA Methods.}
\begin{figure*}[t]
    \centering
    \includegraphics[width=\linewidth]{figures/img/supp_qualitative_results_dexycb.pdf}
    \caption{Qualitative comparison between~\ourname{} and SOTA HPE/HOPE methods across hand-only/hand-object scenarios in DexYCB (``S3" split), in which all the grasping objects are unseen during training.}
    \label{fig:supp_qualitative_results_dexycb}
    \vspace{10mm}
\end{figure*}
\begin{figure*}[t]
    \centering
    \includegraphics[width=\linewidth]{figures/img/supp_qualitative_results_ho3d.pdf}
    \vspace{-7mm}
    \caption{Qualitative comparison between~\ourname{} and SOTA HPE/HOPE methods across hand-only/hand-object scenarios in HO3D. The ground truths are not publicly available.}
    \label{fig:supp_qualitative_results_ho3d}
    \vspace{-3mm}
\end{figure*}
\begin{figure*}[t]
    \centering
    \includegraphics[width=\linewidth]{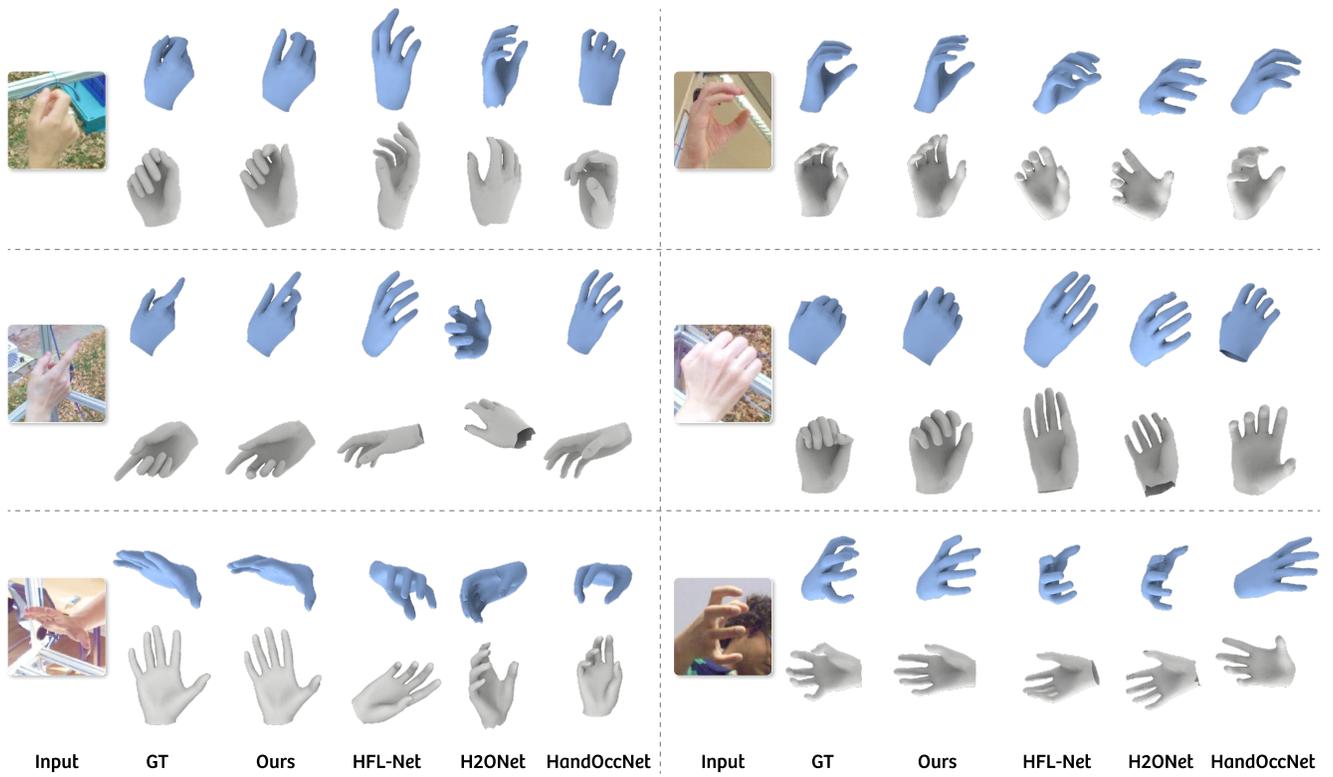}
    \caption{Qualitative comparison between our method and SOTA HPE/HOPE methods on FreiHAND.}
    \label{fig:supp_qualitative_results_freihand}
\end{figure*}
Next, we provide more qualitative comparisons on the DexYCB (\cref{fig:supp_qualitative_results_dexycb}), HO3D (\cref{fig:supp_qualitative_results_ho3d}), and FreiHAND datasets (\cref{fig:supp_qualitative_results_freihand}).

\vspace*{-3mm}
\paragraph{More De-occluded Examples.}
\begin{figure*}[t]
    % \centering
    \includegraphics[width=\linewidth]{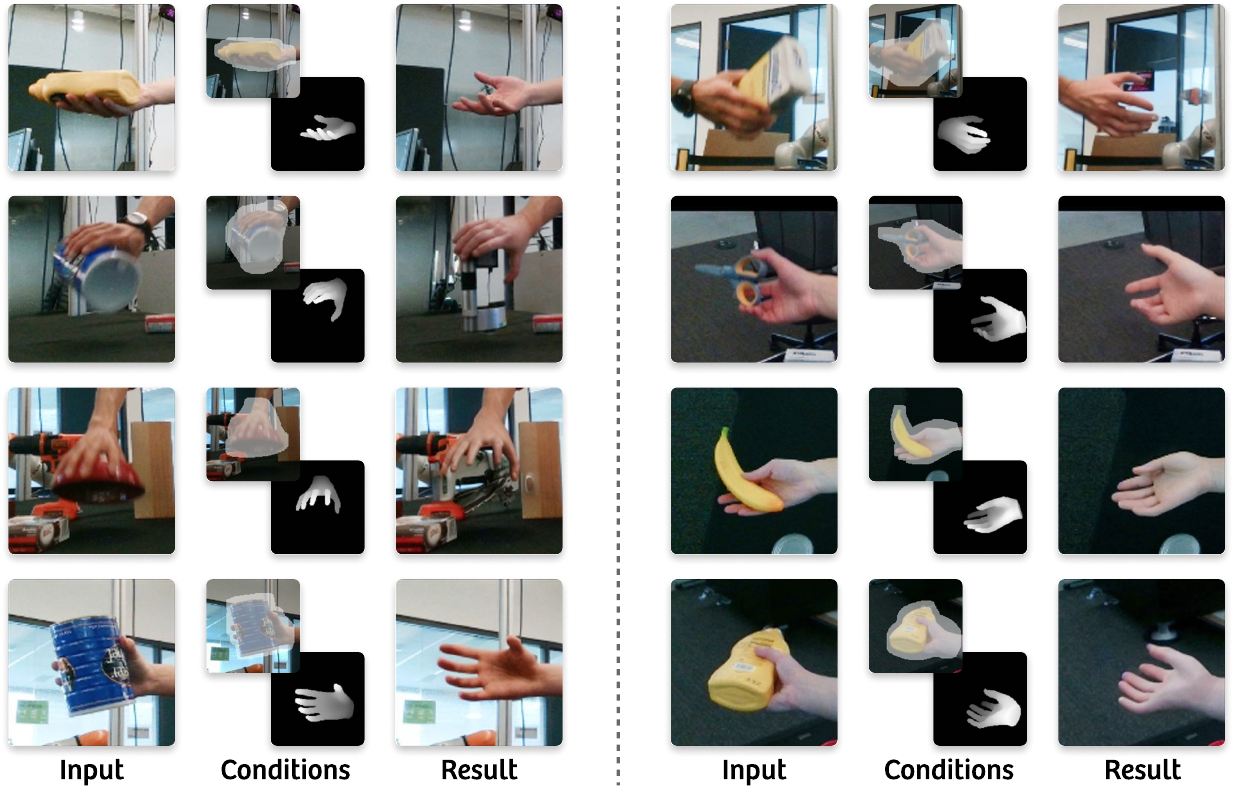}
    \caption{More examples of de-occluded hand images. Note that masks are overlaid on the original image for better visualization, the actual condition for our generative de-occluder is a binary mask.}
    \label{fig:supp_generate_images}
\end{figure*}
Furthermore, we present more de-occluded samples in~\cref{fig:supp_generate_images}.

\subsection{Quantitative Results}
\begin{table*}[t]
    \centering
    \resizebox{\linewidth}{!}{
        \setlength\tabcolsep{2pt}
        \begin{tabular}{c|cccc|cccc|cccc|cccc}
        \toprule
        \multicolumn{1}{c|}{\multirow{2}{*}{\textbf{HPE}}} & \multicolumn{4}{c|}{\textit{Hand-Only Scene}} & \multicolumn{4}{c|}{\textit{Hand-Only $\rightarrow$ Hand-Object Scene}} & \multicolumn{4}{c|}{\textit{All $\rightarrow$ Hand-Only Scene}} & \multicolumn{4}{c}{\textit{All $\rightarrow$ Hand-Object Scene}} \\
        \cmidrule{2-17}
        & J-PE $\downarrow$ & PA-J-PE $\downarrow$ & V-PE $\downarrow$ & PA-V-PE $\downarrow$ & J-PE $\downarrow$ & PA-J-PE $\downarrow$ & V-PE $\downarrow$ & PA-V-PE $\downarrow$ & J-PE $\downarrow$ & PA-J-PE $\downarrow$ & V-PE $\downarrow$ & PA-V-PE $\downarrow$ & J-PE $\downarrow$ & PA-J-PE $\downarrow$ & V-PE $\downarrow$ & PA-V-PE $\downarrow$ \\
        \cmidrule{1-17}
        \cite{park2022handoccnet} & \bslcell{12.98} & \bslcell{5.21} & \bslcell{12.52} & \bslcell{5.02} & 19.60 \worsetext{-6.62} & 7.71 \worsetext{-2.50} & 18.95 \worsetext{-6.43} & 7.42 \worsetext{-2.40} & 13.16 \worsetext{-0.18} & 5.31 \worsetext{-0.10} &  12.70 \worsetext{-0.18} & 5.11 \worsetext{-0.09} &\ignoretext{14.58} & \ignoretext{6.73} & \ignoretext{14.10} & \ignoretext{6.49} \\ \cite{xu2023h2onet} & \bslcell{13.34} & \bslcell{4.69} & \bslcell{13.13} & \bslcell{5.05} & 21.98 \worsetext{-8.64} & 7.13 \worsetext{-2.44} & 21.42 \worsetext{-8.30} & 7.27 \worsetext{-2.22} & 14.14 \worsetext{-0.80} & 4.74  \worsetext{-0.05} & 14.00 \worsetext{-0.87} & 5.35 \worsetext{-0.30} & \ignoretext{15.20} & \ignoretext{6.35} & \ignoretext{15.03} & \ignoretext{6.74} \\
        \cite{zhou2024simple} & \bslcell{14.05} & \bslcell{5.55} & \bslcell{13.51} & \bslcell{5.31} & 18.37 \worsetext{-4.32} & 7.42 \worsetext{-1.87} & 17.54 \worsetext{-4.03} & 6.91 \worsetext{-1.60} & 14.63 \worsetext{-0.58} & 5.62 \worsetext{-0.07} & 13.96 \worsetext{-0.45} & 5.38 \worsetext{-0.07} & \ignoretext{14.88} & \ignoretext{6.74} & \ignoretext{14.21} & \ignoretext{6.45} \\
        \midrule 
        \addlinespace[1mm]
        \midrule
        \multicolumn{1}{c|}{\multirow{2}{*}{\textbf{HOPE}}} & \multicolumn{4}{c|}{\textit{Hand-Object Scene}} & \multicolumn{4}{c|}{\textit{Hand-Object $\rightarrow$ Hand-Only Scene}} & \multicolumn{4}{c|}{\textit{All $\rightarrow$ Hand-Object Scene}} & \multicolumn{4}{c}{\textit{All $\rightarrow$ Hand-Only Scene}} \\
        \cmidrule{2-17}
        & J-PE $\downarrow$ & PA-J-PE $\downarrow$ & V-PE $\downarrow$ & PA-V-PE $\downarrow$ & J-PE $\downarrow$ & PA-J-PE $\downarrow$ & V-PE $\downarrow$ & PA-V-PE $\downarrow$ & J-PE $\downarrow$ & PA-J-PE $\downarrow$ & V-PE $\downarrow$ & PA-V-PE $\downarrow$ & J-PE $\downarrow$ & PA-J-PE $\downarrow$ & V-PE $\downarrow$ & PA-V-PE $\downarrow$ \\
        \cmidrule{1-17}
        \cite{hampali2022keypoint} & \bslcell{17.99} & \bslcell{7.68} & \bslcell{17.57} & \bslcell{7.88} & 25.10 \worsetext{-7.11} & 7.62 \bettertext{+0.06} & 24.40 \worsetext{-6.83} & 7.88 \worsetext{-0.00} & 18.79 \worsetext{-1.00} & 7.77 \worsetext{-0.09} & 18.35 \worsetext{-0.78} & 7.94 \worsetext{-0.06} & \ignoretext{19.75} & \ignoretext{7.59} & \ignoretext{19.26} & \ignoretext{7.98} \\
        \cite{lin2023harmonious} & \bslcell{14.61} & \bslcell{6.56} & \bslcell{14.13} & \bslcell{6.33} & 19.39 \worsetext{-4.78} & 5.96 \bettertext{+0.60} & 18.61 \worsetext{-4.48} & 5.75 \bettertext{+0.58} & 14.77 \worsetext{-0.16} & 6.64 \worsetext{-0.08} & 14.29 \worsetext{-0.16} & 6.41 \worsetext{-0.08} & \ignoretext{13.61} & \ignoretext{5.20} & \ignoretext{13.10} & \ignoretext{5.01} \\        \bottomrule
        \end{tabular}
        }
    \caption{
    Full metrics of Tab.1 in the main paper.
    }
    \label{tab:supp_validation}
\end{table*}
\begin{table*}[!t]
    \centering
    \resizebox{\linewidth}{!}{
        \begin{tabular}{r@{\hskip 6pt}|l@{\hskip 6pt}|cccc|cccc|cccc}
        \toprule
        &\multicolumn{1}{c|}{\multirow{2}{*}{Methods}} & \multicolumn{4}{c|}{\textit{All Scenes}} & \multicolumn{4}{c|}{\textit{Hand-Only Scene}} & \multicolumn{4}{c}{\textit{Hand-Object Scene}} \\
        \cmidrule{3-14}
          & & J-PE $\downarrow$ & PA-J-PE $\downarrow$  & V-PE $\downarrow$ & PA-V-PE $\downarrow$ & J-PE $\downarrow$ & PA-J-PE $\downarrow$  & V-PE $\downarrow$ & PA-V-PE $\downarrow$ & J-PE $\downarrow$ & PA-J-PE $\downarrow$ & V-PE $\downarrow$ & PA-V-PE $\downarrow$ \\
        \midrule
        &HandOccNet~\cite{park2022handoccnet} & 13.04 & 5.85 &  12.61 & 5.65 & 13.42 & 5.39 & 12.95 & 5.20  & 12.79 & 6.15 & 12.39 & 5.95 \\
        &MobRecon~\cite{chen2022mobrecon}  & 14.34 & 6.50 & 13.40 & 5.74 & 14.57 & 5.91 & 13.74 & 5.29  & 14.18 & 6.88 & 13.19 & 6.03 \\
        &H2ONet~\cite{xu2023h2onet}  & 13.89 & \textbf{5.38} &13.56  & 5.52 & 14.10 & \textbf{4.84} & 13.75 & 5.02 & 13.76 & \textbf{5.73} & 13.43 & 5.84 \\
        \multirow{-4}{*}{\rotatebox{90}{HPE}} &SimpleHand~\cite{zhou2024simple}  & 13.66 & 6.02 & 13.14 & 5.78 & 14.48 & 5.67 & 13.95 & 5.46  & 13.13 & 6.24 & 12.62 & 5.99 \\
        \midrule
        &Liu~\etal~\cite{liu2021semi}   & 14.06 & 5.75 & 13.57 & 5.58 & 14.87 & 5.47 & 14.33 & 5.30  & 13.53 & 5.93 & 13.08 & 5.75 \\
        &Keypoint Trans.~\cite{hampali2022keypoint}  & 16.61 & 6.84 & 16.21 & 7.05 & 18.50 & 7.03 & 18.00 & 7.32 & 15.39  & 6.71 & 15.05 & 6.88 \\
        \multirow{-3}{*}{\rotatebox{90}{HOPE}} &HFL-Net~\cite{lin2023harmonious} & 13.02 & 5.58 & \underline{12.58} & \underline{5.39} & \underline{13.41} & 5.19 & \underline{12.92} & \underline{5.00}  & 12.77 & 5.84 & 12.35 & \textbf{5.64} \\
        \midrule
        &H2ONet$^\dagger$ + HFL-Net$^\dagger$   & 13.08 & 5.47 & 12.71 & 5.43 & 13.81 & \underline{4.85} & 13.50 & 5.06 & \underline{12.61} & 5.87 & \underline{12.20} & \underline{5.68} \\
        &H2ONet$^\ddagger$ + HFL-Net$^\ddagger$   & 13.30 & \underline{5.45} & 12.91 & 5.40 & 14.09 & \underline{4.85} & 13.74 & 5.02 & 12.79 & \underline{5.83} & 12.37 & \textbf{5.64} \\
        &HandOccNet$^\dagger$ + HFL-Net$^\dagger$   & 13.32 & 5.73 & 12.87 & 5.54 & 14.40 & 5.50 & 13.89 & 5.30 & 12.63 & 5.89 & 12.22 & 5.69 \\
        &HandOccNet$^\ddagger$ + HFL-Net$^\ddagger$   & 13.43 & 5.71 & 12.97 & 5.51 & 14.41 & 5.49 & 13.90 & 5.30 & 12.80 & 5.85 & 12.38 & 5.65 \\
        \multirow{-5}{*}{\rotatebox{90}{Unified}} &\ourname{} (ours) & \textbf{12.59} & 5.54 & \textbf{12.17} & \textbf{5.36} & \textbf{12.84} & 5.02 & \textbf{12.38} & \textbf{4.85}  & \textbf{12.42} & 5.88 & \textbf{12.03} & 5.69 \\
        \bottomrule
        \end{tabular}
        }
    \caption{Quantitative comparison on DexYCB ``S0'' split. }
    \label{tab:supp_dexycb_s0}
\end{table*}
\begin{table*}[!t]
    \centering
    \resizebox{\linewidth}{!}{
        \begin{tabular}{r@{\hskip 6pt}|l@{\hskip 6pt}|cccc|cccc|cccc}
        \toprule
        &\multicolumn{1}{c|}{\multirow{2}{*}{Methods}} & \multicolumn{4}{c|}{\textit{All Scenes}} & \multicolumn{4}{c|}{\textit{Hand-Only scene}} & \multicolumn{4}{c}{\textit{Hand-Object Scene}} \\
        \cmidrule{3-14}
          & & J-PE $\downarrow$ & PA-J-PE $\downarrow$  & V-PE $\downarrow$ & PA-V-PE $\downarrow$ & J-PE $\downarrow$ & PA-J-PE $\downarrow$  & V-PE $\downarrow$ & PA-V-PE $\downarrow$ & J-PE $\downarrow$ & PA-J-PE $\downarrow$ & V-PE $\downarrow$ & PA-V-PE $\downarrow$ \\
        \midrule
        &HandOccNet~\cite{park2022handoccnet} & 18.33 & 6.95 & 17.70 & 6.71 & 19.70 & 6.01 & 18.95 & 5.81  & 17.57 & 7.47 & 17.02 & 7.21 \\
        &MobRecon~\cite{chen2022mobrecon}  & 18.62 & 7.18 & 17.73 & 6.61 & 19.36 & 6.27 & 18.42 & 5.75  & 18.21 & 7.68 & 17.36 & 7.09 \\
        &H2ONet~\cite{xu2023h2onet}  & 18.40 & \underline{6.40} & 17.90 & 6.57 & 18.92 & \textbf{5.44} & 18.36 & 5.70 & 18.11 & 6.93 & 17.64 & 7.05 \\
        \multirow{-4}{*}{\rotatebox{90}{HPE}} &SimpleHand~\cite{zhou2024simple}  & \underline{17.38} & 6.82 & \underline{16.81} & 6.73 & 18.86 & 6.02 & 18.14 & 5.92  & 16.57 & 7.26 & 16.08 & 7.17 \\
        \midrule
        &Liu~\etal~\cite{liu2021semi}   & 17.82 & 6.46 & 17.19 & \underline{6.25} & 19.12 & 5.89 & 18.36 & 5.69 & 17.10 & \textbf{6.77} & 16.54 & \textbf{6.55} \\
        &Keypoint Trans.~\cite{hampali2022keypoint}  & 21.61 & 8.15 & 21.18 & 8.36 & 22.84 & 7.32 & 22.24 & 7.59 & 20.93 & 8.61 & 20.60 & 8.79 \\
        \multirow{-3}{*}{\rotatebox{90}{HOPE}} &HFL-Net~\cite{lin2023harmonious} & 17.77 & 6.58 & 17.16 & 6.36 & \underline{18.42} & 5.72 & \underline{17.72} & \underline{5.52}  & 17.41 & 7.06 & 16.86 & 6.82 \\
        \midrule
        &H2ONet$^\dagger$ + HFL-Net$^\dagger$   & 17.49 & \textbf{6.36} & 16.94 & \underline{6.25} & 19.24 & 5.50 & 18.60 & 5.59 & \underline{16.54} & \underline{6.83} & \underline{16.02} & \underline{6.61} \\
        &H2ONet$^\ddagger$ + HFL-Net$^\ddagger$   & 17.96 & 6.48 & 17.41 & 6.42 & 18.92 & \underline{5.45} & 18.35 & 5.69 & 17.44 & 7.05 & 16.89 & 6.82 \\
        &HandOccNet$^\dagger$ + HFL-Net$^\dagger$   & 17.84 & 6.53 & 17.22 & 6.31 & 20.18 & 5.95 & 19.39 & 5.75 & 16.55 & 6.85 & 16.03 & 6.62 \\
        &HandOccNet$^\ddagger$ + HFL-Net$^\ddagger$   & 18.63 & 6.73 & 17.99 & 6.50 & 20.72 & 6.11 & 19.91 & 5.91 & 17.48 & 7.06 & 16.93 & 6.83 \\
        \multirow{-5}{*}{\rotatebox{90}{Unified}} &\ourname{} (ours) & \textbf{16.84} & 6.42 & \textbf{16.25} & \textbf{6.20} & \textbf{17.80} & 5.50 & \textbf{17.11} & \textbf{5.30}  & \textbf{16.31} & 6.93 & \textbf{15.79} & 6.70 \\
        \bottomrule
        \end{tabular}
        }
    \caption{Quantitative comparison on DexYCB ``S1'' split. }
    \label{tab:supp_dexycb_s1}
\end{table*}
\paragraph{Additional Results of Tab.~1.} The additional metrics of Tab.~1 in the main paper are provided in~\cref{tab:supp_validation}. Both the metrics before \& after PA show an overall performance degeneration of existing HPE/HOPE models when transferring to apply to the other scenario or testing in the original scenario even after re-training on both scenes.

\vspace*{-3mm}
\paragraph{Comparison on Other Splits of DexYCB.} 
We provide the quantitative results of hand pose estimation on the default ``S0'' split (same distribution for the training and test set) and ``S1'' split with unseen subjects (train/test: 7/2 subjects) of DexYCB in~\cref{tab:supp_dexycb_s0} and~\cref{tab:supp_dexycb_s1}, respectively. Our method achieves the best overall performance, especially in root-relative metrics.

\vspace*{-3mm}
\paragraph{Comparison on HO3D.}
\begin{table*}[t]
    \resizebox{\linewidth}{!}{
        \setlength\tabcolsep{15pt}
        \begin{tabular}{r@{\hskip 6pt}|l@{\hskip 6pt}|cccccc|cc}
        \toprule
          &  \multicolumn{1}{c|}{\multirow{2}{*}{Methods}} & \multicolumn{6}{c|}{\textit{Procrustes Alignment}} & \multicolumn{2}{c}{\textit{Scale-Translation Aligned}} \\
          \cmidrule{3-10}
          & & J-PE $\downarrow$ & J-AUC $\uparrow$ & V-PE $\downarrow$ & V-AUC $\uparrow$ & F@5 $\uparrow$ & F@15 $\uparrow$ & J-PE $\downarrow$ & J-AUC $\uparrow$\\
        \midrule
        &HandOccNet~\cite{park2022handoccnet} & 10.26 & 7.95 & 10.21 & 79.61 & 50.61 & 94.47 & 28.18 & 49.28 \\
        &MobRecon~\cite{chen2022mobrecon}  & 10.47 & 79.14 & 10.76 & 78.54 & 47.57 & 93.59 & 29.36 & 49.36 \\
        &H2ONet~\cite{xu2023h2onet}  & 9.52 & 80.97 & 9.60 & 80.81 & 52.62 & 95.09 & 29.67 & 48.53 \\
        \multirow{-4}{*}{\rotatebox{90}{HPE}} &SimpleHand~\cite{zhou2024simple}  & 11.28 & 77.66 & 11.58 & 77.05 &  45.78 & 91.74 & 28.41 & 49.32 \\
        \midrule
        &Liu~\etal~\cite{liu2021semi}   & 9.46 & 81.12 & 9.39 & 81.25 & 54.93 & 95.64 & 28.44 & 49.79 \\
        &Keypoint Trans.~\cite{hampali2022keypoint}  & 12.00 & 76.24 & 12.18 & 75.83  &  44.71 & 91.60 & 40.00 & 36.36 \\
        \multirow{-3}{*}{\rotatebox{90}{HOPE}} &HFL-Net~\cite{lin2023harmonious} & \underline{9.01} & \underline{82.02} & \underline{8.92} & \underline{82.18} &  \underline{57.01} & \underline{96.19} & 27.97 & 51.33 \\
        \midrule
        &H2ONet$^\dagger$ + HFL-Net$^\dagger$   & 9.49 & 81.04 & 9.43 & 81.16 & 54.54 & 95.54 &  30.60 & 48.93 \\
        &H2ONet$^\ddagger$ + HFL-Net$^\ddagger$  & \textbf{8.97} & \textbf{82.10} & \textbf{8.88} & \textbf{82.26} &  \textbf{57.08} & \textbf{96.22} & 28.00 & 51.44 \\
        &HandOccNet$^\dagger$ + HFL-Net$^\dagger$  & 9.56 & 80.89 & 9.50 & 81.02 & 54.23 & 95.47 & 30.29 & 49.09 \\
        &HandOccNet$^\ddagger$ + HFL-Net$^\ddagger$  & 9.05 & 81.94 & 8.96 & 82.10 & 56.79 & 96.14 & \underline{27.83} & \underline{51.45} \\
        \multirow{-5}{*}{\rotatebox{90}{Unified}} &\ourname{} (ours) & 9.60 & 80.82 & 9.45 & 81.12 &  52.57 & 95.68 & \textbf{25.53} & \textbf{53.70} \\
        \bottomrule
        \end{tabular}
    }
    \caption{Quantitative comparison (\emph{Procrustes Alignment \& Scale-Translation Aligned}) on HO3D.}
    \label{tab:supp_ho3d_v2}
\end{table*}
The remaining hand metrics on HO3D are reported in~\cref{tab:supp_ho3d_v2}. Though HFL-Net~\cite{lin2023harmonious} and the combination of H2ONet + HFL-Net achieve better PA results, our method outperforms them by a large margin in the metrics after scale-translation only alignment~\cite{hampali2020honnotate}, which takes both the global rotation and hand shape into consideration. We emphasize the importance of global rotation, since it better reflects the visualization quality, as indicated by the qualitative comparison results shown in~\cref{fig:supp_qualitative_results_ho3d}.

\subsection{Detailed Analysis on Performance}
\begin{figure*}[t]
    \centering
    \includegraphics[width=0.8\linewidth]{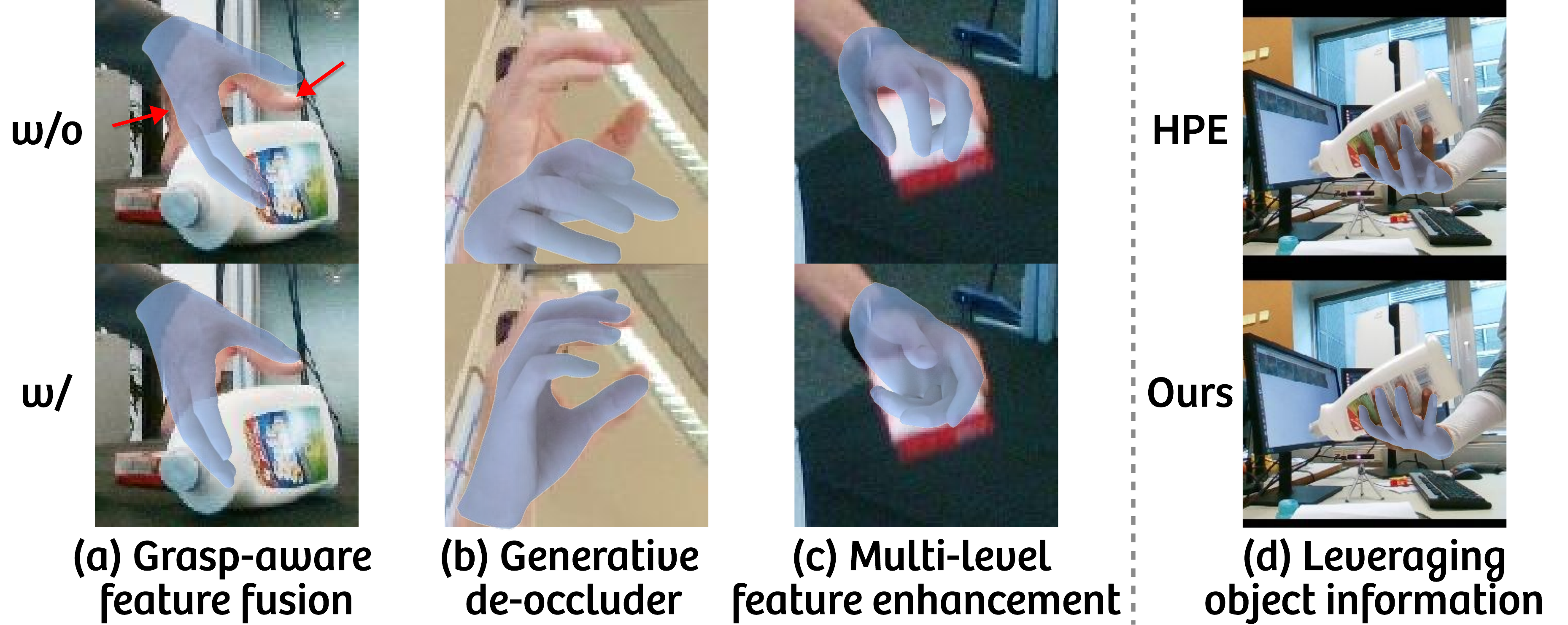}
    % \vspace{-3mm}
    \caption{
    Effects of different designs in our pipeline.
    }
    \label{fig:supp_visual_ablation}
    % \vspace{-7mm}
\end{figure*}
% \wyq{TODO}
In this work, we explore a new setting to address HPE and HOPE at once.
Applying prior SOTA of HPE/HOPE is suboptimal, even re-trained on all scenarios, as they lack specific designs. 
% HPE, HOPE on hand-only and hand-object
For hand-only scenes, HOPE methods are affected by irrelevant object features, even no object is grasped, yet HPE methods may fail for unseen hand poses. 
For hand-object scenes, HOPE methods lack effective designs to handle severe occlusions, while HPE methods do not utilize object information to enhance performance.
Our approach works better in each scene type.
As~\cref{fig:supp_visual_ablation} shows:
(a) when the hand reaches out to grasp an object, our grasp-aware feature fusion reduces the adverse impact of non-grasped object;
(b) for unseen hand poses from FreiHAND, our generated de-occluded images introduce richer hand poses to boost performance; 
(c) our multi-level feature enhancement improves robustness under severe object occlusions; and
(d) when grasping objects, our method surpasses HPE methods by leveraging object information.
These observations are consistent with the quantitative performance in Tab.~2, 5, 6 in the main paper.

\subsection{Additional Ablation Studies}
\begin{table}[t]
    \centering
    \resizebox{\linewidth}{!}{%
    \setlength\tabcolsep{7.5pt}
    \begin{tabular}{l|cc|cc}
        \toprule
        \multicolumn{1}{c|}{\multirow{2}{*}{Control Strength Selection}} & \multicolumn{2}{c|}{\emph{Root-relative}} & \multicolumn{2}{c}{\emph{Procrustes Align.}} \\
        \cmidrule{2-5}
        & J-PE $\downarrow$ & V-PE $\downarrow$ & J-PE $\downarrow$ & V-PE $\downarrow$ \\
        \midrule
        s = 0.4 & 13.76 & 13.30  & 5.85 & 5.65  \\
        s = 0.55 & 13.51 & 13.06  & 5.78 & 5.57  \\
        s = 0.7 & 13.43 & 12.98  & 5.75 & 5.55  \\
        \midrule
        Adaptive Adjustment (ours) & \textbf{13.38} & \textbf{12.92}  & \textbf{5.71} & \textbf{5.52}  \\
         \bottomrule
    \end{tabular}%
    }
    \caption{Quantitative results of our adaptive control strength adjustment~\vs fixed control strengths.}
    \label{tab:supp_ablation_control_strength}
\end{table}
\begin{figure*}[t]
    \centering
    \includegraphics[width=\linewidth]{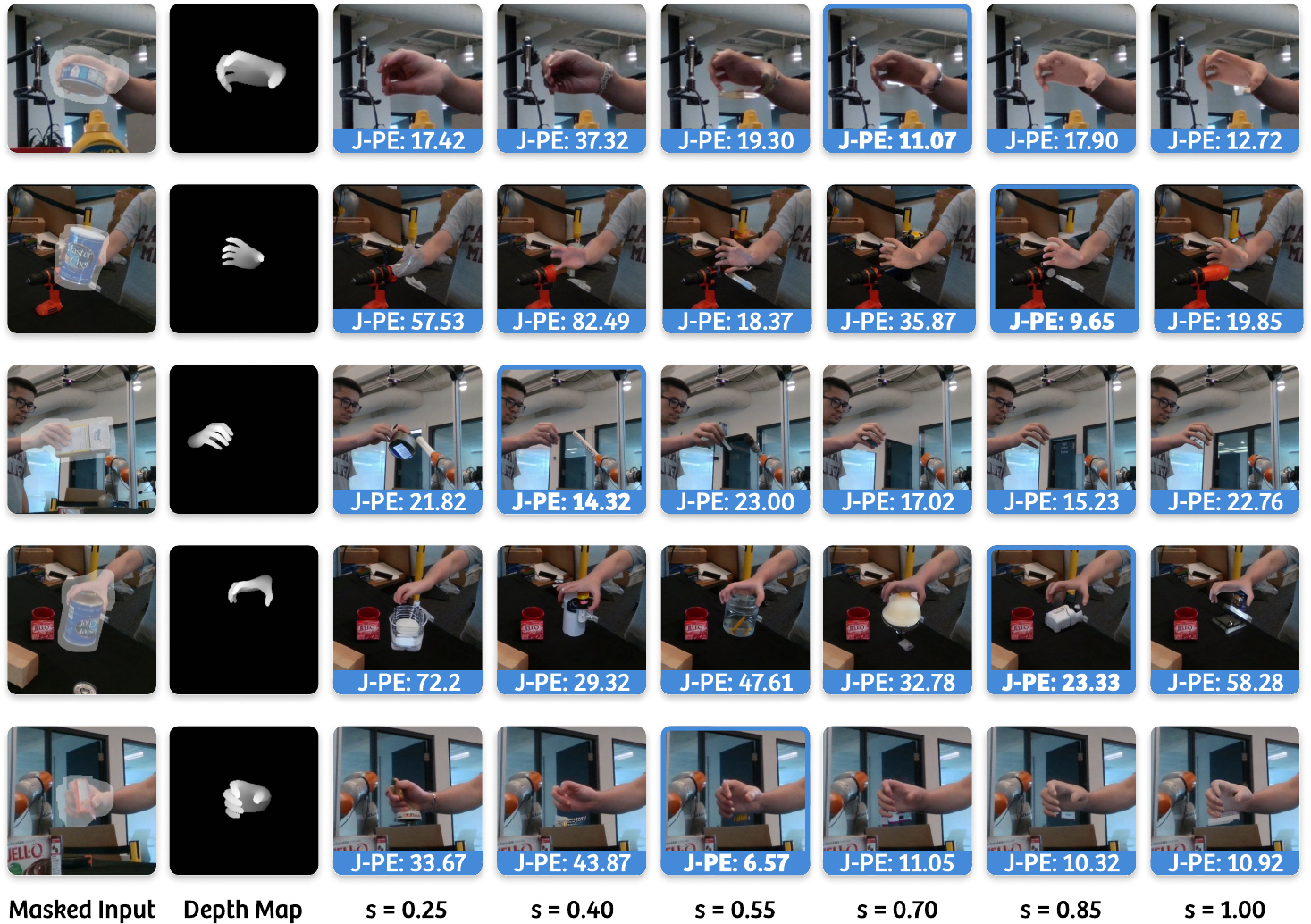}
    \caption{The generated images with varying control strengths. Our adaptive strategy (metrics marked in \textbf{bold}) effectively balances fidelity and consistency.}
    \label{fig:supp_control_strengths_visual}
\end{figure*}
\begin{table}[t]
    \centering
    \resizebox{\linewidth}{!}{%
    \begin{tabular}{l|cc|cc}
        \toprule
        \multirow{2}{*}{Models} & \multicolumn{2}{c|}{\emph{Root-relative}} & \multicolumn{2}{c}{\emph{Procrustes Align.}} \\
        \cmidrule{2-5}
        & J-PE $\downarrow$ & V-PE $\downarrow$ & J-PE $\downarrow$ & V-PE $\downarrow$ \\
        \midrule
        Baseline & \multirow{2}{*}{13.84} & \multirow{2}{*}{13.37} & \multirow{2}{*}{5.79} & \multirow{2}{*}{5.58}  \\ 
        w/ Grasp-aware Feature Fusion & & & & \\
        \midrule
        w/ RHD~\cite{zimmermann2017learning} \& Static Gestures~\cite{staticgestures} & 13.79 & 13.32 & 5.73 & 5.53  \\
        Ours & \textbf{13.03} & \textbf{12.59}  & \textbf{5.59} & \textbf{5.40}  \\
         \bottomrule
    \end{tabular}%
    }
    \caption{Comparison with directly training with synthetic datasets used by~\cite{lu2024handrefiner}.}
    \label{tab:supp_ablation_synthetic_data}
\end{table}
\begin{table}[t]
    \centering
    \resizebox{\linewidth}{!}{%
    \setlength\tabcolsep{10pt}
    \begin{tabular}{c|cc|cc}
        \toprule
        \multirow{2}{*}{$\gamma_{init}$/ $\gamma_{RoI}$ / $\gamma_{MANO}$} & \multicolumn{2}{c|}{\emph{Root-relative}} & \multicolumn{2}{c}{\emph{Procrustes Align.}} \\
        \cmidrule{2-5}
        & J-PE $\downarrow$ & V-PE $\downarrow$ & J-PE $\downarrow$ & V-PE $\downarrow$ \\
        \midrule
        0.001 / 0.001 / 0.005 & 13.17 & 12.72 & 5.61 &  5.41 \\
        0.01 / 0.01 / 0.05 & 13.13 & 12.69 & 5.62 &  5.42 \\
        0.1 / 0.1 / 0.5 (ours) & \textbf{13.03} & \textbf{12.59} & \textbf{5.59} & \textbf{5.40} \\
        1.0 / 1.0 / 5.0 & 13.15 & 12.70 & 5.70 & 5.50 \\
        10.0 / 10.0 / 50.0 & 14.13 & 13.65 & 6.08 & 5.87  \\
         \bottomrule
    \end{tabular}%
    }
    \caption{Effects of various hyperparameters of the multi-level feature constraints.}
    \label{tab:supp_hyperparam}
\end{table}
% As in the main paper, we follow \phil{TODO: XXX citation?} to conduct all the ablation studies presented below on the DexYCB ``S3'' split.
To be consistent with the main paper, we conduct all the ablation studies presented below on DexYCB.
\paragraph{Additional Results of Tab.~7.}
Since the RHD~\cite{zimmermann2017learning} and Static Gestures Dataset~\cite{staticgestures} are utilized to fine-tune the ControlNet~\cite{lu2024handrefiner}, we also conduct an ablation study of pre-training on these synthetic datasets before training on DexYCB, using a network structure identical to our baseline model with the grasp-aware feature fusion module (Row (b) of Tab.~7 in the main paper). 
% to prove the advantages of the occlusion-invariant feature learning strategy. 
As shown in~\cref{tab:supp_ablation_synthetic_data}, directly incorporating synthetic datasets into training leads to a minor improvement, indicating the limitation caused by the domain gap between the synthetic and real-world images.
% possibly due to two reasons: i) The domain gap between the synthetic data and the real-world images. 
% ii) The ground-truth annotations only have 3D joints, lacking supervision on MANO parameters. 
Conversely, our occlusion-invariant feature learning strategy substantially enhances the model performance through the foundational data prior provided by ControlNet~\cite{zhang2023adding} and the multi-level feature enhancement.

\vspace*{-3mm}
\paragraph{Ablation on Adaptive Control Strength Adjustment.}
% to 3-5 lines
% Explain the control strength
Control strength (ranging from 0 to 1) is imposed on the connections between the ControlNet and Stable Diffusion, controlling the extent to which the output is consistent with the control signal. We propose to adaptively adjust its value with MobRecon~\cite{chen2022mobrecon} pre-trained on DexYCB to avoid tedious manual tuning.
% Since the generative de-occluder is trained on synthetic hand images, the generated result might suffer from an unrealistic appearance as the control strength approaches 1, while a rather small control strength might produce an incorrect hand pose. Therefore, we propose to adaptively adjust the value with MobRecon~\cite{chen2022mobrecon} pre-trained on DexYCB to avoid tedious manual tuning.
% As the authors pointed out in their official repository, a control strength of 0.4-0.8 is recommended. 
The default control strength employed in~\cite{lu2024handrefiner} is 0.55.  In our work, we empirically select the candidate control strengths from \{0.25, 0.4, 0.55, 0.7, 0.85, 1.0\}, with a similar number of candidates as in~\cite{lu2024handrefiner}.

To assess the effectiveness of our adaptive control strength adjustment, we compare our model (Row (c) of Tab.~7 in the main paper) with the ones trained with generated samples under fixed control strengths without incorporating the feature enhancement constraints.
As shown in~\cref{tab:supp_ablation_control_strength}, our adaptive strategy achieves the best performance in hand pose estimation compared to several control strengths. 
The samples generated under all candidate control strengths are provided in~\cref{fig:supp_control_strengths_visual}, showing the need to adaptively select control strength for different cases.

\vspace*{-3mm}
\paragraph{Effects of Hyperparameters.}
% The default value of hyperparameter $\alpha$ in Eq.~(11) of the main paper is empirically set to 10. 
The default value of hyperparameter $\alpha$ is empirically set to 10 in Eq.~(11) of the main paper.
This is to ensure a prediction accuracy over 95\%. For the hyperparameters controlling the feature enhancement at three different levels, we evaluate their effects on the hand pose estimation performance in~\cref{tab:supp_hyperparam}. 
Since the MANO-level feature is a late-stage feature employed to directly regress the final hand pose, an adaption layer is deployed to improve the knowledge transfer. We set a larger value for $\gamma_{\scriptscriptstyle{M\!A\!N\!O}}$ to aim to strongly enforce this feature adaptation process. In our experiments, the values for $\gamma_{init}$, $\gamma_{\scriptscriptstyle{RoI}}$, and $\gamma_{\scriptscriptstyle{M\!A\!N\!O}}$ are set to 0.1, 0.1, and 0.5, respectively.

\subsection{Computational Cost and Efficiency}
The training time of our model is 3 days for DexYCB (376k samples) and 12 hours for HO3D (66k samples), respectively, on eight NVidia RTX 2080Ti GPUs.

\begin{table*}[t]
\centering
    \resizebox{\linewidth}{!}{
        \setlength\tabcolsep{2.5pt}
        \begin{tabular}{c|cccccccccc}
        \toprule
        Methods & HandOccNet~\cite{park2022handoccnet} & MobRecon~\cite{chen2022mobrecon} &
        H2ONet~\cite{xu2023h2onet} &
        SimpleHand~\cite{zhou2024simple} &
        Liu~\etal~\cite{liu2021semi} & 
        Keypoint Trans.~\cite{hampali2022keypoint} & 
        HFL-Net~\cite{lin2023harmonious} & 
        H2ONet + HFL-Net &
        HandOccNet + HFL-Net &
        Ours \\ 
        \midrule
        FPS     & 48   & 78  &  62  & 41  & 51  & 33  & 43 & 36 & 30 & 44 \\
        \midrule
          FLOPs     & 15.48G   & 0.46G   &  0.74G    & 9.96G  & 39.44G  & 12.66G & 10.01G & 0.77G / 10.04G & 15.51G / 10.04G & 10.04G \\
        \midrule
         \# Param.     & 37.22M   & 8.23M   &  25.88M    & 48.89M  & 34.48M  & 52.79M  & 46.08M & 72.26M & 83.60M & 46.38M \\
        \bottomrule
        \end{tabular}}
    \caption{Efficiency comparison with previous methods. Note that FLOPs for the ``A+B'' methods depend on the predicted grasping status, therefore reported as ``FLOPs of (classifier + A) / FLOPs of (classifier + B)''.}
    \label{tab:supp_efficiency}
\end{table*}
\cref{tab:supp_efficiency} reports the inference speed (FPS, tested on a single NVidia RTX 2080Ti GPU), FLOPs, and number of parameters of various models. Thanks to the lightweight object switcher in \ourname{}, \ourname{} has similar inference efficiency and model complexity as HFL-Net~\cite{lin2023harmonious}. Compared to other SOTA models, \ourname{} has a moderate model size and running speeds, enabling real-time applications.

% \subsection{Detailed Analysis on Feature Enhancement}
% placeholder. Show some features before/after enhancement, and show their MSE Loss.

\section{Implementation Details}
\label{sec:supp_implementation_details}

\paragraph{Scene Division.}
% In the training set, the numbers of samples for the two scenes are 145,051/231,323 for DexYCB, and 5,595/60,439 for HO3D,respectively. For the test set, the numbers are 29,912/46,448 for DexYCB and 2,971/8,553 for HO3D. Note that FreiHAND cannot be split in this manner due to the lack of object annotations.
\begin{table*}
    \centering
    \resizebox{\linewidth}{!}{%
    \setlength\tabcolsep{15pt}
    \begin{tabular}{c|ccc|ccc}
        \toprule
        \multirow{2}{*}{Datasets (splits)} & \multicolumn{3}{c|}{\emph{Training Set}} & \multicolumn{3}{c}{\emph{Test Set}} \\
        \cmidrule{2-7}
        & All Scenes & Hand-Only Scene & Hand-Object Scene & All Scenes & Hand-Only Scene & Hand-Object Scene \\
        \midrule
        DexYCB ``S0" & 401,507 & 153,210 & 248,297 & 78,768 & 30,848 & 47,920 \\
        DexYCB ``S1" & 351,943 & 138,775 & 213,168 & 104,128 & 36,912 & 67,216 \\
        DexYCB ``S3" & 376,374 & 145,051 & 231,323 & 76,360 & 29,912 & 46,448 \\
        HO3D & 66,034 & 5,595 & 60,439 & 11,524 & 2,971 & 8,553 \\
        FreiHAND & 130,240 & N/A & N/A & 3,960 & N/A & N/A \\
         \bottomrule
    \end{tabular}%
    }
    \caption{Number of samples in hand-only/hand-object scenes for different datasets (splits).}
    \label{tab:supp_tab_scene_splits}
\end{table*}
Following~\cite{xu2024handbooster}, the thresholds for RRE and RTE in grasping label preparation are 5$\degree$ and 10mm, respectively. An image is categorized into the hand-only scenes, if determined as non-grasping, otherwise hand-object scenes. The numbers of samples in the two scenes are shown in~\cref{tab:supp_tab_scene_splits}. Note that although FreiHAND~\cite{zimmermann2019freihand} contains a small number of images interacting with objects in both training and test sets, it cannot be divided due to the lack of object annotations.

\vspace*{-3mm}
\paragraph{Generative De-occluder.}
We adopt the officially-released pre-trained weights from~\cite{lu2024handrefiner}, which fine-tunes ControlNet with synthetic hand images~\cite{zimmermann2017learning, staticgestures}. The hand-object mask is obtained by applying dilation on the render mask of the 3D hand and object to ensure the hand-object region is covered for repainting. Then,
% and perform the inference to obtain the de-occluded images.
% The ControlNet takes an image with 512x512 resolution as input. 
% to accommodate the input size to the diffusion model, 
we crop the original input image in the training set centered on the hand-object region and resize it to $512\!\times\!512$. The hand-object image and the hand-object mask are fed into the inpainting Stable Diffusion model, conditioned by the hand depth map.
Besides, we adopt the positive prompt ``a hand grasping gesture, indoor, in the lab'' for image generation from the two laboratory benchmarks~\cite{chao2021dexycb, hampali2020honnotate}, and the negative prompt is similar to the one in~\cite{lu2024handrefiner}. 
During inference, the number of reverse steps for DDIM is set to 50 by default.

\vspace*{-3mm}
\paragraph{Network Structure.}
(i) \textbf{Backbone}: Following~\cite{lin2023harmonious}, we adopt ResNet50~\cite{he2016deep} as the backbone to extract features from the input image, in which a dual stream structure is adopted to relieve the competition between hand features and object features.
(ii) \textbf{Hand Encoder}: The hand encoder takes $\mathbf{F}^{OH}$ as input, first using an hourglass network~\cite{newell2016human} to regress a feature map and the heatmap of 2D hand joints. Then, they are fused via a convolution layer and an element-wise addition, followed by four residual blocks to yield a 1024-dimensional vector.
(iii) \textbf{MANO Decoder}: It consists of two fully connected layers to predict the hand pose and shape parameters of the MANO model from the feature produced by the hand encoder.
(iv) \textbf{Object Decoder}: Following~\cite{lin2023harmonious}, the feature after RoIAlign from the hand branch is fused with the one from the object branch through a cross-attention layer, to enhance the object feature learning. The fused feature is then forwarded through six convolutional layers to predict the 2D projections of the 3D object corner keypoints and corresponding confidence. In testing, the object pose is computed by the Perspective-n-Point (PnP) algorithm~\cite{lepetit2009ep} using the correspondence between the predicted 2D and the original 3D keypoints on the object mesh.
% More details about the network structure can be found in~\cite{lin2023harmonious}.

\vspace*{-3mm}
\paragraph{Training Details.}
Following~\cite{lin2023harmonious}, we perform data augmentation on the training samples, including random scaling ($\pm20\%$), rotating ($\pm180\degree$), translating ($\pm10\%$), and color jittering ($\pm50\%$).
Our training process consists of two stages. In the first stage, the de-occluded images are incorporated into training without the feature enhancement loss for 30 epochs to first adapt the model to the domain of the generated data. 
In the second stage, the network is additionally supervised by the enhancement constraints between the image pairs for another 40 epochs under the same setting.

\section{Limitations and Future Work}
\label{sec:supp_limitations}

\paragraph{Limitations.} 
Though we are able to predict the grasping status of unseen objects, the performance of their pose estimation tends to degrade when the object shape/appearance varies largely, due to the limited object categories in the training data. 
Besides, despite being provided in most existing public benchmarks, the object annotations are lacking in certain datasets, limiting the applicability of our approach as they are required for scene division and inpainting masks.

\vspace*{-3mm}
\paragraph{Future Work.} 
To improve the model's generalizability towards unseen objects, a promising direction is to utilize the knowledge prior from the various vision foundation models~\cite{radford2021learning, kirillov2023segment, liu2024visual}, which demonstrated remarkable performance in zero-shot scenarios. Another approach that we are considering for improving the model's generalizability is to train on large-scale synthetic data by leveraging diffusion models~\cite{xu2024handbooster, vaswani2017attention} or large language models~\cite{wen2024foundationpose}.
{
    \small
    \bibliographystyle{ieeenat_fullname}
    \bibliography{review}
}